\documentclass[reqno,11pt]{amsart}

\usepackage{fancyhdr}
\usepackage[margin=1in]{geometry}
\usepackage{amsmath,amssymb,amsthm}
\usepackage{graphicx}
\usepackage{url}
\usepackage{color}
\usepackage{chngcntr}
\usepackage{caption}
\counterwithin{figure}{section}
\numberwithin{equation}{section}

\newcommand{\XXX}{\mathcal{X}}
\newcommand{\PPP}{\mathcal{P}}
\newcommand{\RR}{\mathbb{R}}
\newcommand{\CC}{\mathbb{C}}
\newcommand{\ZZ}{\mathbb{Z}}
\newcommand{\GGG}{\mathcal{G}}
\newcommand{\tr}{\operatorname{tr}}
\newcommand{\rank}{\operatorname{rank}}

\begin{document}

\title{Non-unique games over compact groups and orientation estimation in cryo-EM}

\author{Afonso S. Bandeira}
\address{Program in Applied and Computational Mathematics (PACM), Princeton University, Princeton, NJ 08544, USA}
\email{ajsb@math.princeton.edu}
\thanks{
This research was partially supported by Award Number R01GM090200 from the NIGMS, FA9550-12-1-0317 and FA9550-13-1-0076 from AFOSR, LTR DTD 06-05-2012 from the Simons Foundation, and the Moore Foundation Data-Driven Discovery Investigator Award.  
}

\author{Yutong Chen}
\address{Program in Applied and Computational Mathematics (PACM), Princeton University, Princeton, NJ 08544, USA}
\email{yutong@math.princeton.edu}

\author{Amit Singer}
\address{Department of Mathematics and PACM, Princeton University, Princeton, NJ 08544, USA}
\email{amits@math.princeton.edu}

\begin{abstract}
Let $\GGG$ be a compact group and let $f_{ij} \in L^2(\GGG)$.
We define the \emph{Non-Unique Games} (NUG) problem as finding $g_1,\dots,g_n \in \GGG$ to minimize 
$\sum_{i,j=1}^n f_{ij} \left( g_i g_j^{-1}\right)$.
We devise a relaxation of the NUG problem to a \emph{semidefinite program} (SDP) by taking the Fourier transform of $f_{ij}$ over $\GGG$, which can then be solved efficiently.
The NUG framework can be seen as a generalization of the \emph{little Grothendieck} problem over the orthogonal group and the \emph{Unique Games} problem and includes many practically relevant problems, such as the \emph{maximum likelihood estimator} to registering bandlimited functions over the unit sphere in $d$-dimensions and orientation estimation in cryo-Electron Microscopy.
\end{abstract}

\maketitle

\section{Introduction}\label{sec:intro}

We consider problems of the following form
\begin{equation}\label{eq:NUG}
\begin{aligned}
& \underset{g_1,\ldots,g_n}{\operatorname{minimize}} && \sum_{i,j=1}^n f_{ij} \left( g_i g_j^{-1} \right)\\
& \text{subject to} && g_i \in \GGG,
\end{aligned}
\end{equation}
where $\GGG$ is a compact group and $f_{ij} : \GGG \rightarrow \mathbb{R}$ are suitable functions. 
We will refer to such problems as a \emph{Non-Unique Game} (NUG) problem over $\GGG$.

Note that the solution to the NUG problem is not unique.
If $g_1,\ldots,g_n$ is a solution to \eqref{eq:NUG}, then so is $g_1 g , \ldots , g_n g$ for any $g \in \GGG$.
That is, we can solve \eqref{eq:NUG} up to a global shift $g \in \GGG$.

Many inverse problems can be solved as instances of~\eqref{eq:NUG}.
A simple example is angular synchronization~\cite{ASinger_2011_angsync,Bandeira_rankrecoveryangsynch}, where one is tasked with estimating angles $\{\theta_i\}_i$ from information about their offsets $\theta_i-\theta_j \mod 2\pi$.
The problem of estimating the angles can then be formulated as an optimization problem depending on the offsets, and thus be written in the form of~\eqref{eq:NUG}.
In general, many inverse problems, where the goal is to estimate multiple group elements from information about group offsets, can be formulated as~\eqref{eq:NUG}.

One of the simplest instances of~\eqref{eq:NUG} is the \texttt{Max-Cut} problem, where the objective is to partition the vertices of a graph as to maximize the number of edges (the \emph{cut}) between the two sets.
In this case, $\GGG\cong \ZZ_2$, the group of two elements $\{\pm1\}$, and $f_{ij}$ is zero if $(i,j)$ is not an edge of the graph and
\[
 f_{ij}(1) = 0 \text{ and } f_{ij}(-1) = -1,
\]
 if $(i,j)$ is an edge.
In fact, we take a semidefinite programming based approach towards~\eqref{eq:NUG} that is inspired by --- and can be seen as a generalization of--- the semidefinite relaxation for the \texttt{Max-Cut} problem by Goemans and Williamson~\cite{MXGoemans_DPWilliamson_1995}.

Another important source of inspiration is the semidefinite relaxation of \texttt{Max-2-Lin}($\ZZ_L$), proposed in~\cite{Charikar_Makarychev_UG}, for the \emph{Unique Games} problem, a central problem in theoretical computer science~\cite{SKhot_2002,SKhot_2010}.
Given integers $n$ and $L$, an Unique-Games instance is a system of linear equations over $\ZZ_L$ on $n$ variables $\left\{ x_i  \right\}_{i=1}^n$.
Each equation constraints the difference of two variables.
More precisely, for each $(i,j)$ in a subset of the pairs, we associate a constraint
\[
x_i - x_j = b_{ij} \mod L.
\]
The objective is then to find $\left\{ x_i  \right\}_{i=1}^n$ in $\ZZ_L$ that satisfy as many equations as possible.
This can be easily described within our framework by taking, for each constraint,
\[
f_{ij}(g) = -\delta_{g\equiv b_{ij}},
\]
and $f_{ij} = 0$ for pairs not corresponding to constraints.
The term ``unique'' derives from the fact that the constraints have this special structure where the offset can only take one value to satisfy the constraint, and all other values have the same score. This motivated our choice of nomecluture for the framework treated in this paper. The semidefinite relaxation for the unique games problem proposed in~\cite{Charikar_Makarychev_UG} was investigated in~\cite{Bandeira_MultireferenceAlignment} in the context of the signal alignment problem, where the $f_{ij}$ are not forced to have a special structure (but $\GGG\cong \ZZ_L$). The NUG framework presented in this paper can be seen as a generalization of the approach in~\cite{Bandeira_MultireferenceAlignment} to other compact groups $\GGG$.

Besides the signal alignment treated in~\cite{Bandeira_MultireferenceAlignment} the semidefinite relaxation to the NUG problem we develop coincides with other effective relaxations. When $\GGG\cong \ZZ_2$ it coincides with the semidefinite relaxations for \texttt{Max-Cut}~\cite{MXGoemans_DPWilliamson_1995}, little Grothendieck problem over $\ZZ_2$~\cite{NAlon_ANaor_2006,Nesterov_quadprogram1}, recovery in the stochastic block model~\cite{Abbe_SBMExact,Bandeira_Laplacian}, and Synchronization over $\ZZ_2$~\cite{Abbe_Z2Synch,Bandeira_Laplacian,Cucuringu_Z2Synch}.
When $\GGG\cong SO(2)$ and the functions $f_{ij}$ are linear with respect to the representation $\rho_1:SO(2)\to \CC$ given by the $\rho_1(\theta) = e^{i\theta}$, it coincides with the semidefinite relaxation for angular synchronization~\cite{ASinger_2011_angsync}. Similarly, when $\GGG\cong O(d)$ and the functions are linear with respect to the natural $d$-dimensional representation, then the NUG problem essentially coincides with the little Grothendieck problem over the orthogonal group~\cite{Bandeira_LittleGrothendieckOd,Naor_etal_NCGI}. Other examples include the shape matching problem in computer graphics for which $\GGG$ is a permutation group (see~\cite{Huang_Guibas_Graphics,Chen_Huang_Guibas_Graphics}).

\subsection{Orientation estimation in cryo-Electron Microscopy}

A particularly important application of this framework is the orientation estimation problem in cryo-Electron Microscopy~\cite{ASinger_YShkolnisky_commonlines}.

\begin{figure}[h]
\begin{center}
\includegraphics[width = 0.35\textwidth]{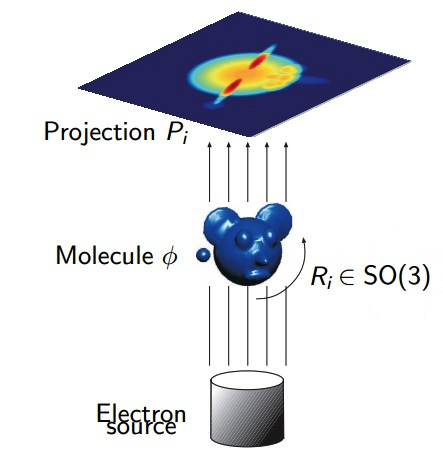}
\quad 
\includegraphics[width = 0.45\textwidth]{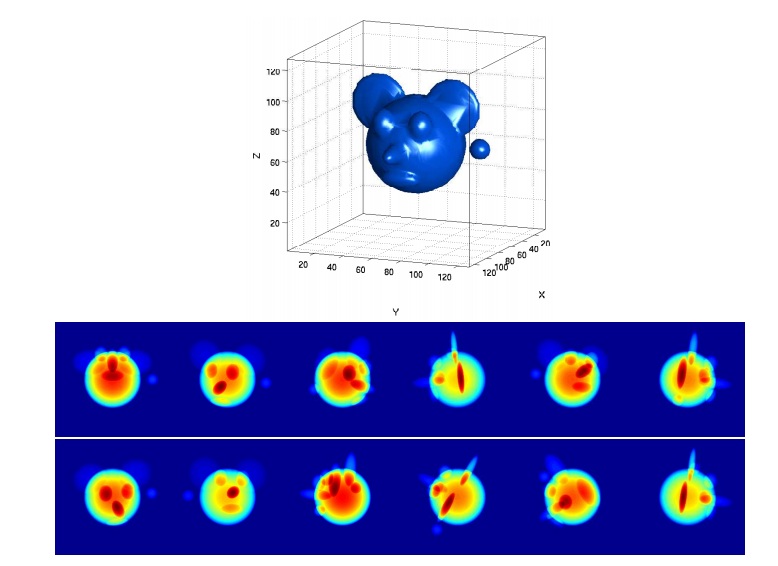}
\caption{Illustration of the cryo-EM imaging process: A molecule is imaged after being frozen at a random (unknown) rotation and a tomographic 2-dimensional projection is captured. Given a number of tomographic projections taken at unknown rotations, we are interested in determining such rotations with the objective of reconstructing the molecule density. Images courtesy of Amit Singer and Yoel Shkolnisky~\cite{ASinger_YShkolnisky_commonlines}.}
\label{fig:cryoEM_intro}
\end{center}
\end{figure}

Cryo-EM is a technique used to determine the 3-dimensional structure of biological macromolecules.
The molecules are rapidly frozen in a thin layer of ice and imaged with an electron microscope, which gives noisy 2-dimensional projections.
One of the main difficulties with this imaging process is that these molecules are imaged at different unknown orientations in the sheet of ice and each molecule can only be imaged once (due to the destructive nature of the imaging process).
More precisely, each measurement consists of a tomographic projection of a rotated (by an unknown rotation) copy of the molecule.
The task is then to reconstruct the molecule density from many such noisy measurements.
In Section~\ref{sec:MRA}, we describe how this problem can be formulated in the form~\eqref{eq:NUG}.

\section{Multireference Alignment}\label{sec:MRA}

In classical linear inverse problems, one is tasked with recovering an unknown element $x\in \XXX$ from a noisy measurement of the form $\PPP(x) + \epsilon$, where $\epsilon$ represents the measurement error and $\PPP$ is a linear observation operator.
There are, however, many problems where an additional difficulty is present; one class of such problems includes non-linear inverse problems in which an unknown transformation acts on $x$ prior to the linear measurement.
Specifically, let $\XXX$ be a vector space and $\GGG$ be a group acting on $\XXX$.
Suppose we have $n$ measurements of the form
\begin{equation}\label{eq:obsmodel}
y_i = \PPP ( g_i \circ x ) + \epsilon_i, \quad i = 1,\dots,n
\end{equation}
where
\begin{itemize}
\item $x$ is a fixed but unknown element of $\XXX$,
\item $g_1,\dots,g_n$ are unknown elements of $\GGG$,
\item $\circ$ is the action of $\GGG$ on $\XXX$,
\item $\PPP : \XXX \to Y$ is a linear operator,
\item $Y$ is the  (finite-dimensional) measurement space,
\item $\epsilon_i$'s are independent noise terms.
\end{itemize}
If the $g_i$'s were known, then the task of recovering $x$ would reduce to a classical linear inverse problem, for which many effective techniques exist.
For this reason, we focus on the problem of estimating $g_1,\ldots,g_n$.

There are several common approaches for inverse problems of the form~\eqref{eq:obsmodel}.
One is motivated by the observation that estimating $x$ knowing the $g_i$'s and estimating the $g_i$'s knowing $x$ are both considerably easier tasks. This suggests a alternating minimization approach where each estimation is updated iteratively.
Besides a lack of theoretical guarantees and a tendency to stall at local optima, these kind of approaches usually start with an initial guess for $x$ and this can introduce model bias (c.f. the experiment in Figure~\ref{fig:RefvsMRA_modelbias}).
Another approach, which we refer to as pairwise comparisons~\cite{ASinger_2011_angsync}, consists in determining, from pairs of observations $(y_i,y_j)$, the most likely value for $g_ig_j^{-1}$.
Although the problem of estimating the $g_i$'s from these pairwise guesses is fairly well-understood~\cite{ASinger_2011_angsync,Bandeira_Singer_Spielman_OdCheeger,ASinger_ZZhao_YShkolnisky_RHadani_cryo} enjoying efficient algorithms and performance guarantees, this method suffers from loss of information as not all of the information of the problem is captured in this most likely value for $g_ig_j^{-1}$ and thus this approach tends to fail at low signal-to-noise-ratio. 

In contrast, the \emph{Maximum Likelihood Estimator} (MLE) leverages all information and enjoys many theoretical guarantees.
Assuming that the $\epsilon_i$'s are i.i.d.\ Gaussian, the MLE for the observation model \eqref{eq:obsmodel} is given by the following optimization problem:
\begin{equation}\label{eq:MRA}
\begin{aligned}
& \underset{g_1,\ldots,g_n,x}{\operatorname{minimize}} && \sum_{i=1}^n \left\| y_i - \PPP ( g_i \circ x ) \right\|_2^2\\
& \text{subject to} && g_i \in \GGG\\
&&& x \in \XXX
\end{aligned}
\end{equation}
We refer to~\eqref{eq:MRA} as the \emph{Multireference Alignment} (MRA) problem.
Unfortunately, the exponentially large search space and nonconvex nature of~\eqref{eq:MRA} often render it computationaly intractable.
However, for several problems of interest, we formulate~\eqref{eq:MRA} as an instance of an NUG for which we develop efficient approximations.

\subsection{Registration of signals on the sphere}

Consider the problem of estimating a bandlimited signal on the circle $x : S^1\to \CC$ from noisy rotated copies of it.
In this problem, $\XXX = \operatorname{span} \left\{ e^{\mathrm{i} k \theta} \right\}_{k=-t}^t$ is the space of bandlimited functions up to degree $t$ on $S^1$, $\GGG = SO(2)$ and the group action is
\[
g \circ x = \sum_{k = -t}^t \alpha_k e^{\mathrm{i} k (\theta - \theta_g)},
\]
where $x \in \XXX$ and we identified $g \in SO(2)$ with $\theta_g \in [0,2\pi]$.

The measurements are of the form
\[
y_i := \PPP ( g_i \circ x ) + \epsilon_i, \quad i = 1,\ldots,n
\]
where
\begin{itemize}
\item $x \in \XXX$,
\item $g_i \in SO(2)$,
\item $\PPP : \XXX \rightarrow \CC^L$ samples the function at $L$ equally spaced points in $S^1$,
\item $\epsilon_i \sim \mathcal{N}(0,\sigma^2 I_{L \times L})$ ($i = 1,\ldots,n$) are independent Gaussians.
\end{itemize}
Our objective is to estimate $g_1,\ldots,g_n$ and $x$. Since estimating $x$ knowing the group elements $g_i$ is considerably easier, we will focus on estimating $g_1,\dots,g_n$. As shown below, this will essentially reduce to the problem of aligning (or registering) the observations $y_1,\dots,y_n$.

\begin{figure}[h!]
\centering
\includegraphics[width = 0.24\textwidth]{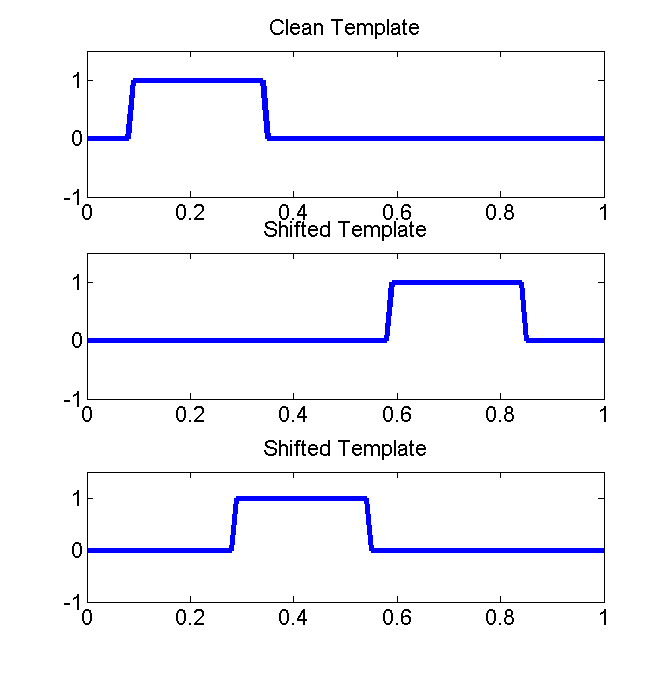}
\includegraphics[width = 0.24\textwidth]{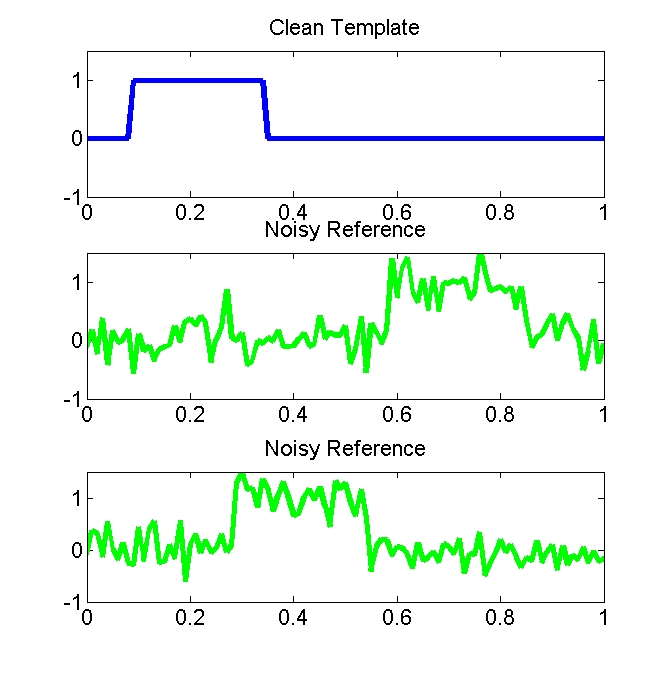}
\includegraphics[width = 0.24\textwidth]{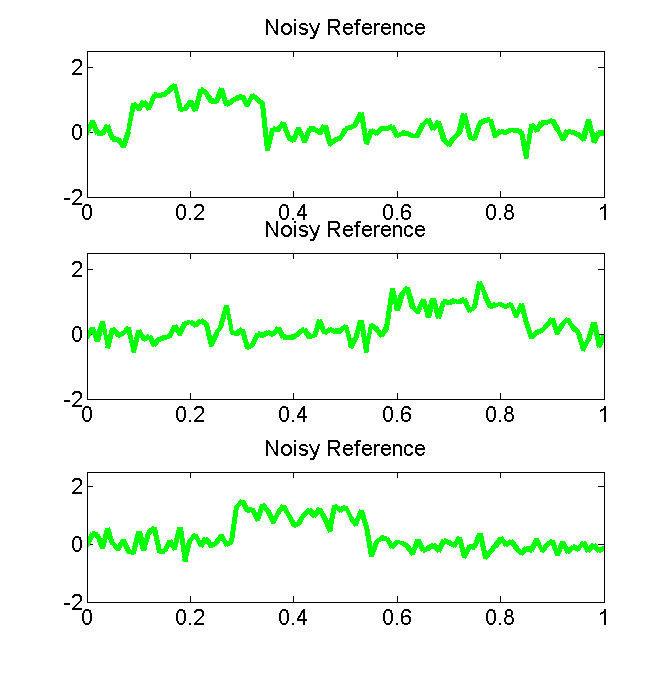}
\includegraphics[width = 0.24\textwidth]{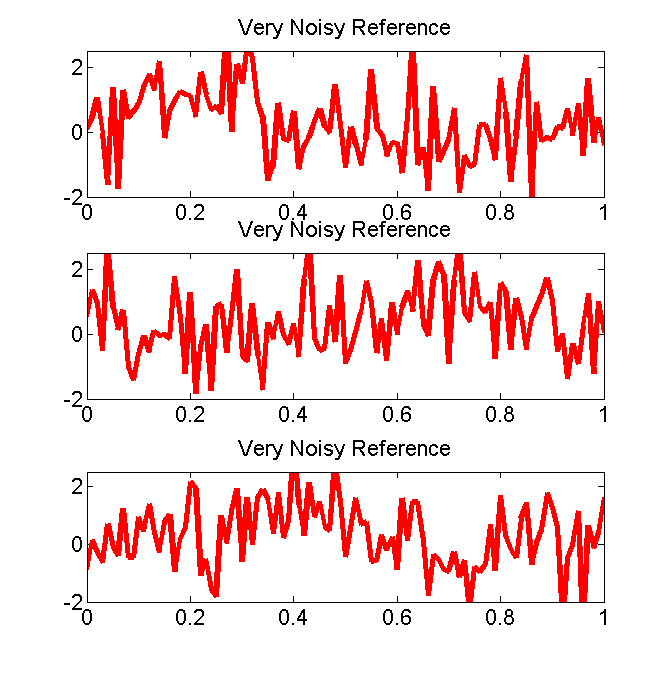}
\caption{
Illustration of the registration problem in $S^1$.
The first column consists of a noiseless example, the second column represents an instance for which the template is known and \emph{matched filtering} is effective.
However, in the examples we are interested in the template is unknown (last two columns) rendering the problem significantly harder.
}
\label{fig:reg1d}
\end{figure}

In absence of noise, the problem of finding the $g_i$'s is trivial (cf. first column of Figure~\ref{fig:reg1d}).
With noise, if $x$ is known (as it is in some applications), then the problem of determining the $g_i$'s can be solved by \emph{matched filtering} (cf. second column of Figure~\ref{fig:reg1d}).
However, $x$ is unknown in general.
This, together with the high levels of noise, render the problem significantly more difficult (cf. last two columns of Figure~\ref{fig:reg1d}).

\begin{figure}[h!]
\centering
\includegraphics[width = 0.8\textwidth]{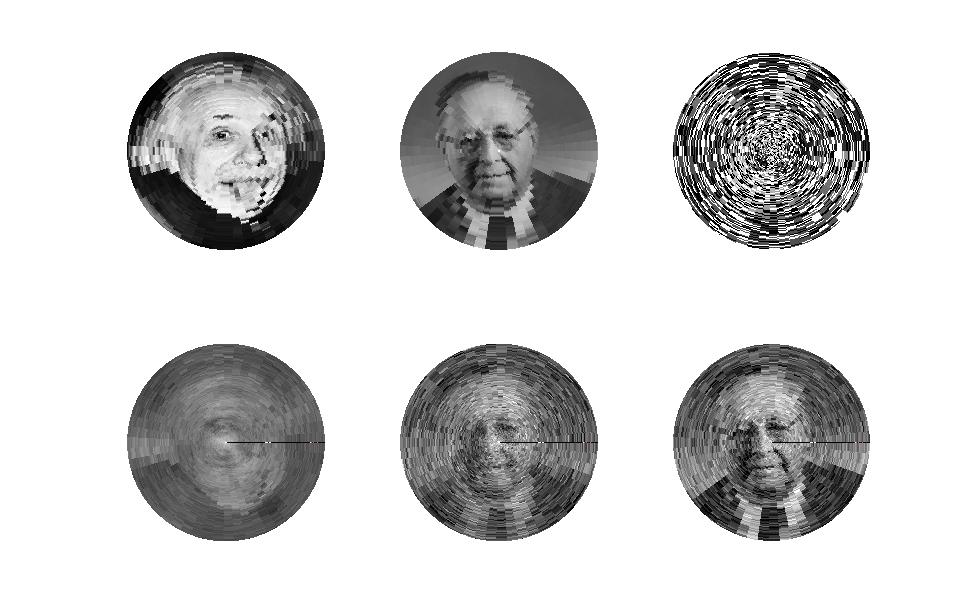}
\caption{
A simple experiment to illustrate the model bias phenomenon: given a picture of the mathematician Hermann Weyl (second picture of the top row) we generate many images consisting of random rotations (we considered a discretization of the rotations of the plane) of the image with added gaussian noise.
An example of one such measurements is the third image in the first row.
We then proceeded to align these images to a reference consisting of a famous image of Albert Einstein (often used in the model bias discussions).
After alignment, an estimator of the original image was constructed by averaging the aligned measurements.
The result, first image on second row, is clearly closer to the image of Einstein than the one of Weyl, illustrating the model bias issue.
On the other hand, the method proposed in~\cite{Bandeira_MultireferenceAlignment} and generalized here produces the second image of the second row, which shows no signs of suffering from model bias.
As a benchmark, we also include the reconstruction obtained by an oracle that is given the true rotations (third image in the second row).
}
\label{fig:RefvsMRA_modelbias}
\end{figure}

In fact, in the high noise regime, if one attempts to perform \emph{matched filtering} with a reference signal $y_j$ that is not the true template $x$ (as this is unknown), then there is a high risk of \emph{model bias}: the reconstructed signal $x$ tends to capture characteristics of the reference $y_j$ that are not present in the actual original signal $x$ (see Figure~\ref{fig:RefvsMRA_modelbias}).
This issue is well known among the biomedical imaging community (see~\cite{Cohen_ModelBias} for a recent discussion).
As Figure~\ref{fig:RefvsMRA_modelbias} suggests, the methods treated in this paper do not suffer from model bias as they do not use any information besides the data itself.

We now define the problem of registration in $d$-dimensions in general.
$\XXX = \operatorname{span} \{ p_k \}_{k \in \mathcal{A}_t}$ is the space of bandlimited functions up to degree $t$ on $S^d$ where the $p_k$'s are orthonormal polynomials on $S^d$, $\mathcal{A}_t$ indexes all $p_k$ up to degree $t$ and $\GGG = SO(d+1)$.

The measurements are of the form
\begin{equation}\label{eq:obsmodel_regd}
y_i := \PPP ( g_i \circ x ) + \epsilon_i, \quad i = 1,\ldots,n
\end{equation}
where
\begin{itemize}
\item $x \in \XXX$,
\item $g_i \in SO(d+1)$,
\item $\PPP : \XXX \rightarrow \CC^L$ samples the function on $L$ points in $S^d$,
\item $\epsilon_i \sim \mathcal{N}(0,\sigma^2 I_{L \times L})$ ($i = 1,\ldots,n$) are independent Gaussians.
\end{itemize}
Again, our objective is to estimate $g_1,\ldots,g_n$ and $x$.


\begin{figure}[h!]
\centering
\includegraphics[width = 0.45\textwidth]{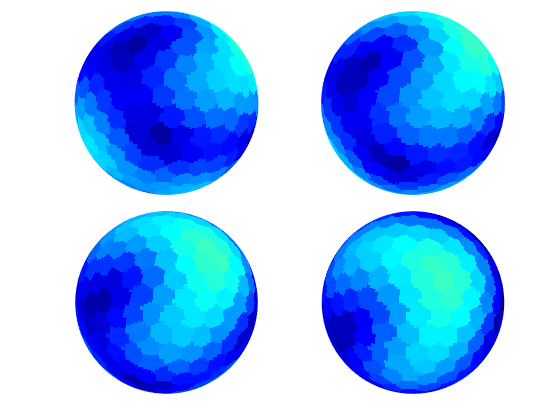}
\includegraphics[width = 0.45\textwidth]{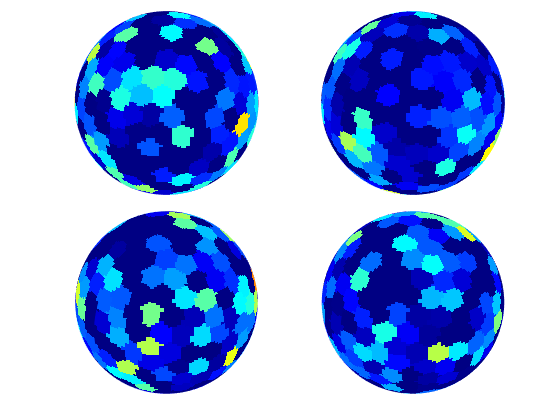}
\caption{
An illustraction of the registration in $2$-dimensions. The left four spheres provide examples of clean signals $y_i$ and the right four spheres are of noisy observations. 
Note that the images are generated using a quantization of the sphere.
}
\label{fig:reg2d}
\end{figure}


The MRA solution for registration in $d$-dimensions is given by
\begin{equation}\label{eq:MRA_regd}
\begin{aligned}
& \underset{g_1,\ldots,g_n,x}{\operatorname{minimize}} && \sum_{i=1}^n \left\| y_i - \PPP ( g_i \circ x ) \right\|_2^2\\
& \text{subject to} && g_i \in SO(d+1)\\
&&& x \in \XXX
\end{aligned}
\end{equation}

Let us remove $x$ from \eqref{eq:MRA_regd}.
Let
\[
\begin{aligned}
\mathcal{Q} : \CC^L &\rightarrow \XXX\\
y_i &\mapsto \sum_{l = 1}^L y_i(l) \operatorname{sinc} (\omega - \omega_l),
\end{aligned}
\]
where $\left\{ \omega_l \right\} \subset S^d$ are the points sampled by $\PPP$ and $\operatorname{sinc}$ is the multidimensional $\operatorname{sinc}$ function.
We define the group action $\cdot$ on the $y_i$ through its action on $\XXX$ by
\[
g^{-1} \cdot y_i := \PPP \left(  g^{-1} \circ \mathcal{Q}(y_i) \right).
\]

The group action of $\GGG = SO(d+1)$ preserves the norm $\| \cdot \|_2$.
Then \eqref{eq:MRA_regd} is equivalent to
\begin{equation}\label{eq:MRA_regd_v2}
\begin{aligned}
& \underset{g_1,\ldots,g_n,x}{\operatorname{minimize}} && \sum_{i=1}^n \left\| g_i^{-1} \cdot y_i - \PPP ( x ) \right\|_2^2\\
& \text{subject to} && g_i \in SO(d+1)\\
&&& x \in \XXX.
\end{aligned}
\end{equation}

Recall that $\XXX$ is the space of bandlimited functions.
If $\PPP$ samples $x$ at sufficiently many well spread-out points (i.e. take a large $L$), then $\| g_i^{-1} y_i \|_2 = \| y_i \|_2$ and $\| \PPP ( x ) \|_2$ can be easily estimated and thus considered approximately fixed.
We approximate \eqref{eq:MRA_regd_v2} with
\begin{equation}\label{eq:MRA_regd_v3}
\begin{aligned}
& \underset{g_1,\ldots,g_n,x}{\operatorname{maximize}} && \sum_{i=1}^n \left\langle g_i^{-1} \cdot y_i , \PPP ( x ) \right\rangle\\
& \text{subject to} && g_i \in SO(d+1).
\end{aligned}
\end{equation}

For fixed $g_i$'s, the maximizing $x$ must satisfy $\PPP ( x ) = \frac{1}{n} \sum_{i=1}^n g_i^{-1} \cdot y_i$.
Then \eqref{eq:MRA_regd_v3} is equivalent to
\begin{equation}\label{eq:MRA_regd_v4}
\begin{aligned}
& \underset{g_1,\ldots,g_n}{\operatorname{maximize}} && \sum_{i=1}^n \left\langle g_i^{-1} \cdot y_i , \frac{1}{n} \sum_{j=1}^n g_j^{-1} \cdot y_j \right\rangle\\
& \text{subject to} && g_i \in SO(d+1).
\end{aligned}
\end{equation}

We simplify the objective function in \eqref{eq:MRA_regd_v4} and get the equivalent problem
\begin{equation}\label{eq:MRA_regd_v5}
\begin{aligned}
& \underset{g_1,\ldots,g_n}{\operatorname{maximize}} && \sum_{i,j=1}^n \left\langle y_i , g_i g_j^{-1} \cdot y_j \right\rangle\\
& \text{subject to} && g_i \in SO(d+1).
\end{aligned}
\end{equation}

Again, we use the observation that $\| y_i \|_2$'s are approximately fixed.
Then, \eqref{eq:MRA_regd_v5} is equivalent to
\begin{equation}\label{eq:MRA_regd_approx}
\begin{aligned}
&\underset{g_1,\ldots,g_n}{\operatorname{minimize}} && \sum_{i,j=1}^n \left\| y_i - g_i g_j^{-1} \cdot y_j \right\|_2^2\\
& \text{subject to} && g_i \in SO(d+1).
\end{aligned}
\end{equation}

In summary, \eqref{eq:MRA_regd} can be approximated by \eqref{eq:MRA_regd_approx}, which is an instance of \eqref{eq:NUG}.

\subsection{Orientation estimation in cryo-EM}

%

The task here is to reconstruct the molecule density from many such measurements (see the second column of Figure~\ref{fig:cryoEM_intro} for an idealized density and measurement dataset).
The linear inverse problem of recovering the molecule density given the rotations fits in the framework of classical computerized tomography for which effective methods exist.
Thus, we focus on the non-linear inverse problem of estimating the unknown rotations and the underlying density. 

\begin{figure}[h!]
\centering
\includegraphics[width = 0.24\textwidth]{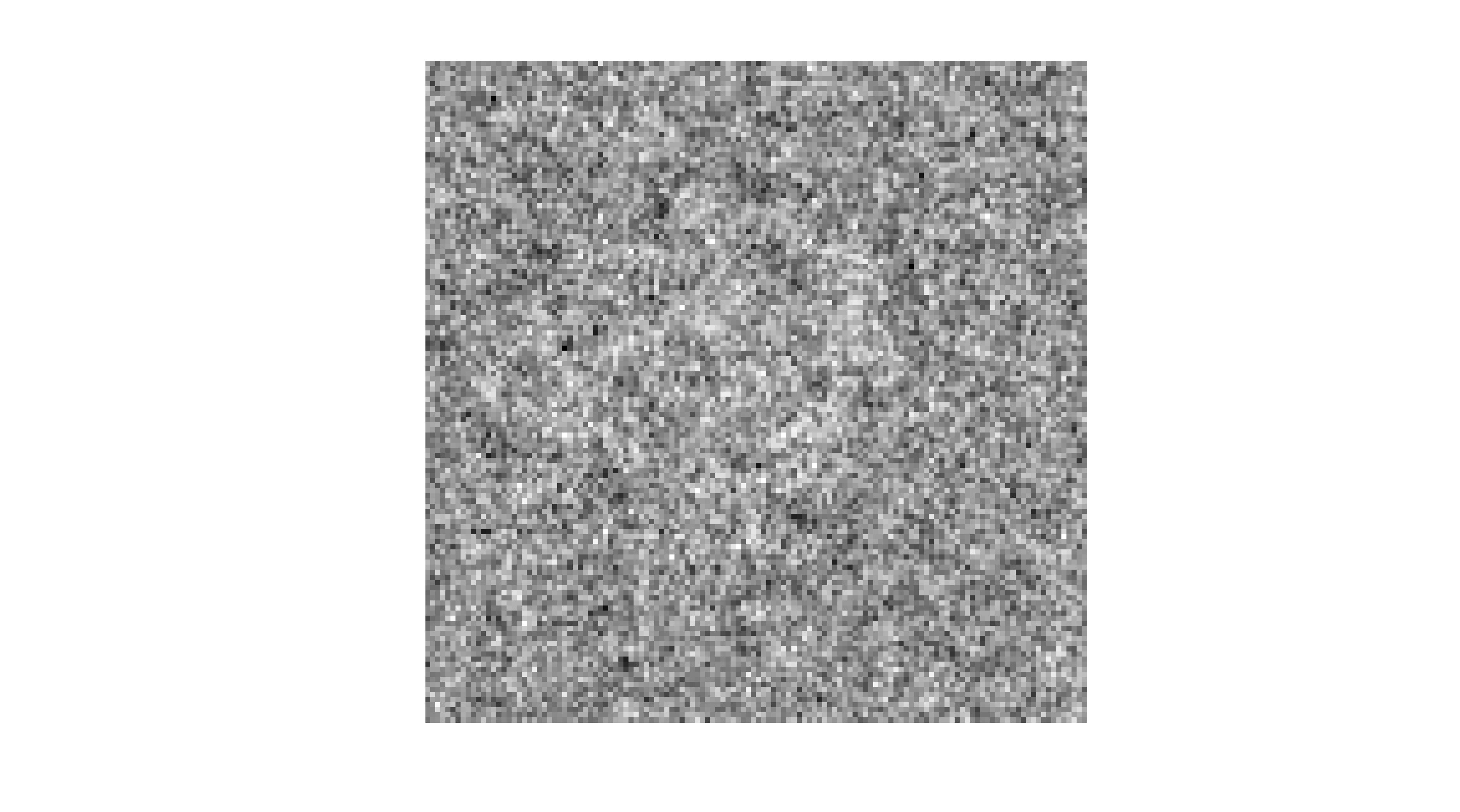}
\includegraphics[width = 0.24\textwidth]{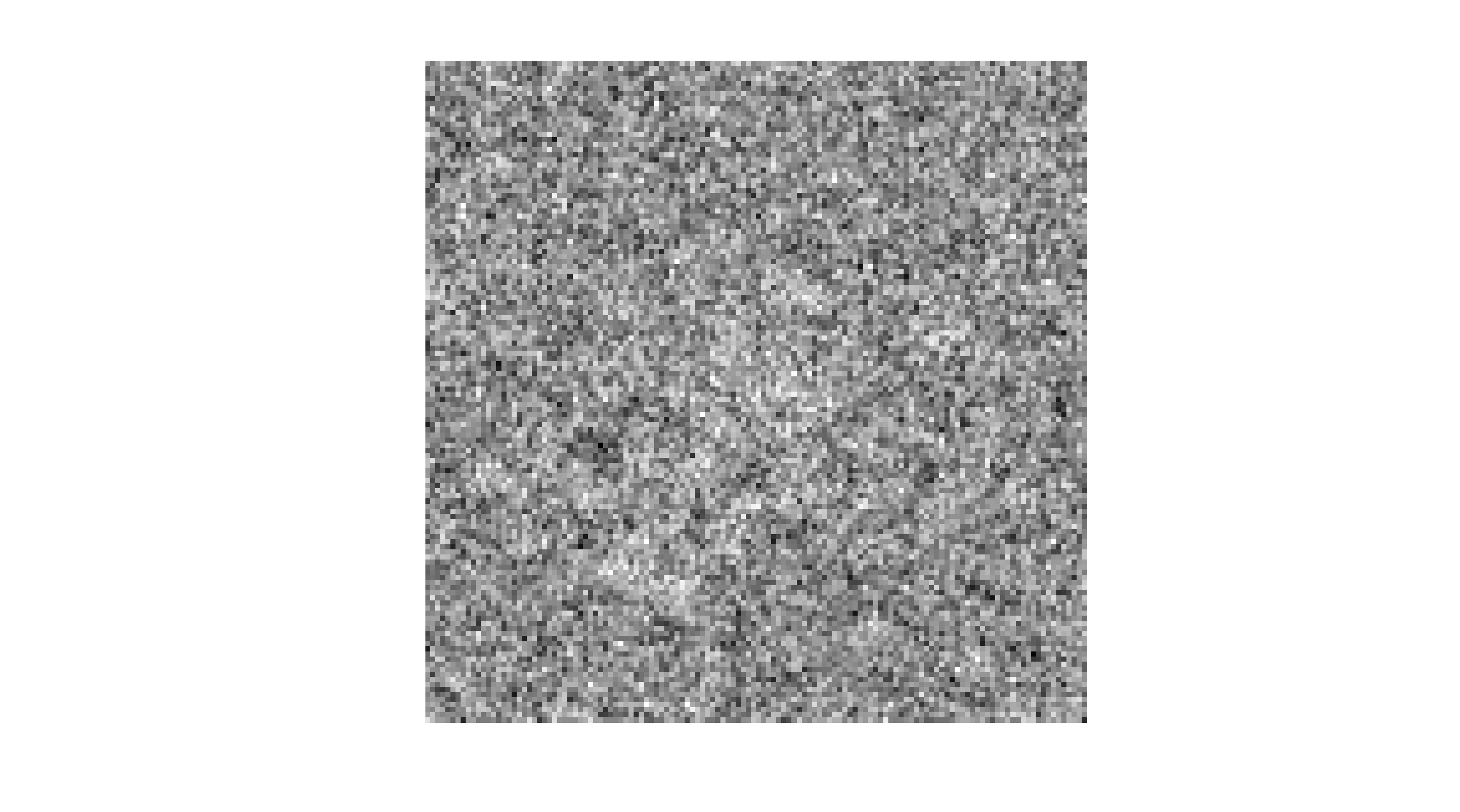}
\includegraphics[width = 0.24\textwidth]{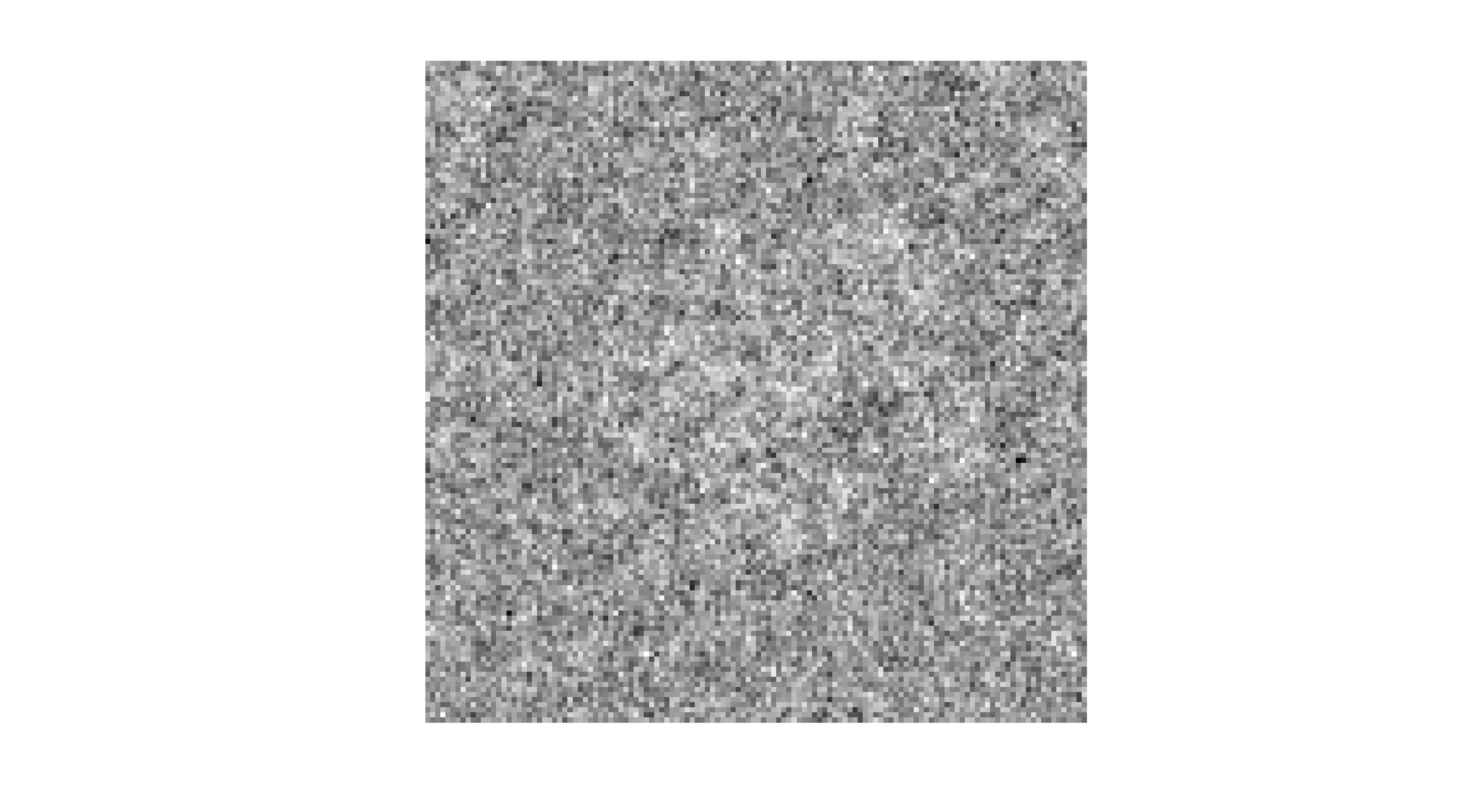}
\includegraphics[width = 0.24\textwidth]{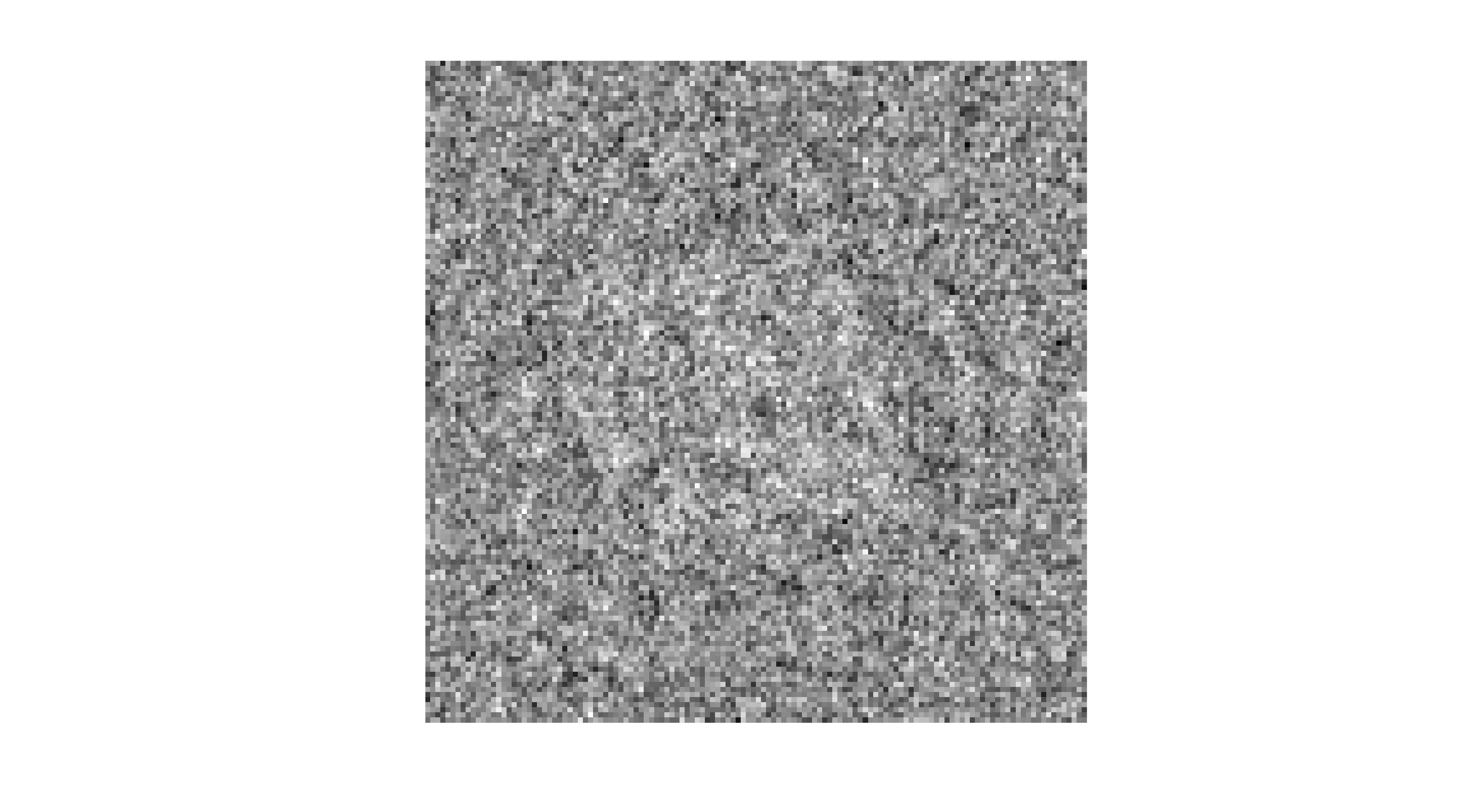}
\caption{Sample images from the E. coli 50S ribosomal subunit, generously provided by Fred Sigworth at the Yale Medical School.}
\label{fig:noisy_image}
\end{figure}

An added difficulty is the high level of noise in the images.
In fact, it is already non-trivial to distinguish whether a molecule is present in an image or if the image consists only of noise (see Figure~\ref{fig:noisy_image} for a subset of an experimental dataset).
On the other hand, these datasets consist of many projection images which renders reconstruction possible.

We formulate the problem of orientation estimation in cryo-EM.
Let $\XXX$ to be the space of bandlimited functions that are also essentially compactly supported in $\RR^3$ and $\GGG = SO(3)$.
The measurements are of the form
\begin{equation}\label{eq:obsmodel_cryoEM}
I_i(x,y) := \PPP ( R_i \circ \phi ) + \epsilon_i, \quad i = 1,\ldots,n
\end{equation}
\begin{itemize}
\item $\phi \in \XXX$,
\item $R_i \in SO(3)$,
\item $\PPP ( \phi )$ samples $\int_{-\infty}^{\infty} \phi(x,y,z) dz$ ($\PPP$ is called the discrete X-ray transform),
\item $\epsilon_i$'s are i.i.d Gaussians representing noise.
\end{itemize}
Our objective is to find $g_1,\ldots,g_n$ and $\phi$.

The operator $\PPP$ in the orientation estimation problem is different than in the registration problem.
Specifically, $\PPP$ is a composition of tomographic projection and sampling, thereby rendering the MLE more involved.
To write the MLE solution for the orientaiton estimation problem, we will use the \emph{Fourier slice theorem}~\cite{book:Natterer}.

\begin{figure}[h!]
\centering
\includegraphics[width = 0.6\textwidth]{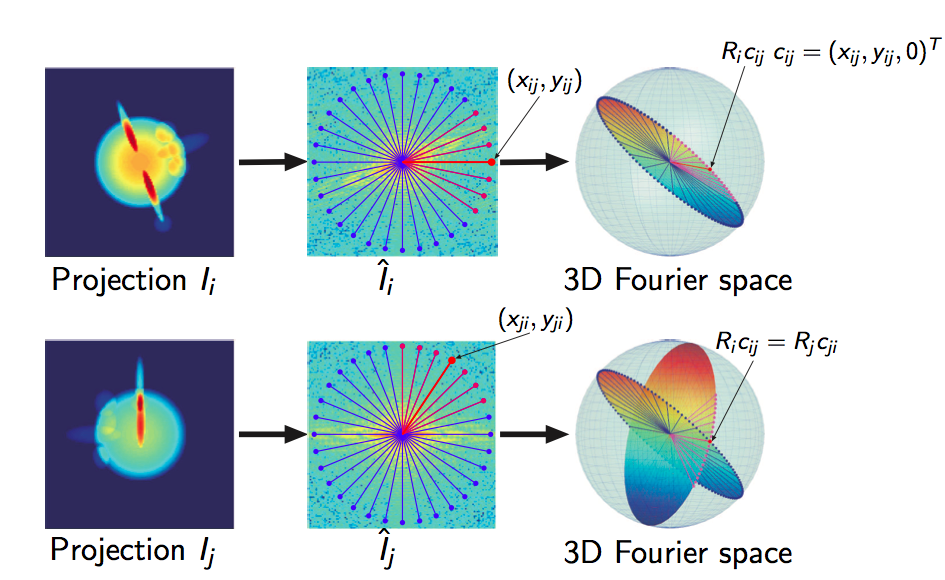}
\caption{An illustration of the use of the Fourier slice theorem and the common lines approach to the orientation estimation problem in cryo-EM.}
\label{fig:commonlines_slicethm}
\end{figure}

The \emph{Fourier slice theorem} states that the $2$-dimensional Fourier transform of a tomographic projection of a molecule density $\phi$ coincides with the restriction to a plane normal to the projection direction, a slice, of the $3$-dimensional Fourier transform of the density $\phi$.
See Figure~\ref{fig:commonlines_slicethm}.

Let $\hat{I}_i(r,\theta)$ be the Fourier transform of $I_i$ in polar coordinates.
We embed $\hat{I}_i$ and $\hat{I}_j$ in $\RR^3$, and apply $g_i^{-1}$ and $g_j^{-1}$ to $\hat{I}_i$ and $\hat{I}_j$, respectively.
Then, the directions of the lines of intersection on $\hat{I}_i$ and $\hat{I}_j$ are given, respectively, by unit vectors
\begin{align}
c_{ij} \left( g_i g_j^{-1} \right) =& \frac{g_i ( g_i^{-1} \cdot \vec{e}_3 \times g_j^{-1} \cdot \vec{e}_3 )}{\left\| g_i ( g_i^{-1} \cdot \vec{e}_3 \times g_j^{-1} \cdot \vec{e}_3 \right\|_2} = \frac{ \vec{e}_3 \times g_i g_j^{-1} \cdot \vec{e}_3 }{\| \vec{e}_3 \times g_i g_j^{-1} \cdot \vec{e}_3 \|},\label{eq:cryoEM_cline_i}\\
c_{ji} \left( g_i g_j^{-1} \right) =& \frac{g_j ( g_j^{-1} \cdot \vec{e}_3 \times g_i^{-1} \cdot \vec{e}_3 )}{\left\| g_j ( g_j^{-1} \cdot \vec{e}_3 \times g_i^{-1} \cdot \vec{e}_3 ) ) \right\|_2} =  \frac{\vec{e}_3 \times \left( g_i g_j^{-1} \right)^{-1} \cdot \vec{e}_3}{\| \vec{e}_3 \times \left( g_i g_j^{-1} \right)^{-1} \cdot \vec{e}_3 \|}.\label{eq:cryoEM_cline_j}
\end{align}
where $\vec{e}_3 := (0,0,1)^T$.
See \cite{ASinger_YShkolnisky_commonlines} for details.

We seek the MRA solution on the lines of intersection.
\begin{equation}\label{eq:MRA_cryoEM}
\begin{aligned}
& \underset{g_1,\ldots,g_n}{\operatorname{minimize}} && \sum_{i,j = 1}^n \left\| \hat{I}_i \left( \cdot , c_{ij} \left( g_i g_j^{-1} \right) \right)- \hat{I}_i \left( \cdot , c_{ji} \left( g_i g_j^{-1} \right) \right) \right\|_2^2\\
& \text{subject to} && g_i \in SO(3),
\end{aligned}
\end{equation}
and \eqref{eq:MRA_cryoEM} is an instance of \eqref{eq:NUG}.

Note that for $n = 2$ images, there is always a degree of freedom along the line of intersection.
In otherwords, we cannot recover the true orientation between $\hat{I}_1$ and $\hat{I}_2$.
However, for $n \geq 3$, this degree of freedom is eliminated. In general, the measurement system suffers from a handness ambuiguity on the reconstruction (see, for example,~\cite{ASinger_YShkolnisky_commonlines}), this will be discussed in detail in a future version of this manuscript. It is also worth mentioning several important references in the context of angular reconstitution~\cite{vanheel_angularreconstitution,Vainshtein_Goncharov}.

\section{Linearization via Fourier expansion}\label{sec:NUG_SDP}

Let us consider the objective function in the general form
\begin{equation}\label{eq:NUG_objective}
\sum_{i,j=1}^n f_{ij} \left( g_i g_j^{-1} \right).
\end{equation}
Note that each $f_{ij}$ in \eqref{eq:NUG_objective} can be nonlinear and nonconvex.
However, since $\GGG$ is compact (and since $f_{ij} \in L^2(\GGG)$), we can expand, each $f_{ij}$ in \emph{Fourier series}.
More precisely, given the unitary irreducible representations $\{ \rho_k \}$ of $\GGG$, we can write
\begin{equation}
\begin{aligned}
f_{ij} \left( g_i g_j^{-1} \right) =& \sum_{k = 0}^{\infty} d_k \tr \left[ \hat{f}_{ij}(k) \rho_k \left( g_i g_j^{-1} \right) \right]\\
=& \sum_{k = 0}^{\infty} d_k \tr \left[ \hat{f}_{ij}(k) \rho_k(g_i) \rho_k^*(g_j) \right],
\end{aligned}
\end{equation}
where $\hat{f}_{ij}(k)$ are the \emph{Fourier coefficients} of $f_{ij}$ and can be computed from $f_{ij}$ via the \emph{Fourier transform}
\begin{equation}
\begin{aligned}
\hat{f}_{ij}(k) :=& \int_\GGG f_{ij}(g) \rho_k(g^{-1}) dg\\
=& \int_\GGG f_{ij}(g) \rho_k^*(g) dg.
\end{aligned}
\end{equation}
Above, $dg$ denotes the \emph{Haar measure} on $\GGG$ and $d_k$ the dimension of the representation $\rho_k$.

We express the objective function \eqref{eq:NUG_objective} as
\[
\begin{aligned}
\sum_{i,j=1}^n f_{ij} \left( g_i g_j^{-1} \right) =& \sum_{i,j=1}^n \sum_{k = 0}^\infty d_k \tr \left[ \hat{f}_{ij}(k) \rho_k(g_i) \rho_k^*(g_j) \right]\\
=& \sum_{k = 0}^\infty \sum_{i,j=1}^n d_k \tr \left[ \hat{f}_{ij}(k) \rho_k(g_i) \rho_k^*(g_j) \right],
\end{aligned}
\]
which is linear in $ \rho_k(g_i) \rho_k^*(g_j)$.
This motivates writing~\eqref{eq:NUG} as linear optimization over the variables
\[
X^{(k)} := \begin{bmatrix} \rho_k(g_1) \\ \vdots \\ \rho_k(g_n) \end{bmatrix} \begin{bmatrix} \rho_k(g_1) \\ \vdots \\ \rho_k(g_n) \end{bmatrix}^*.
\]
In other words,
\[
 \sum_{i,j=1}^n f_{ij} \left( g_i g_j^{-1} \right) = \sum_{k=0}^\infty \tr\left[ C^{(k)}X^{(k)} \right],
\]
where the coefficient matrices are given by 
\[
C^{(k)} := d_k \begin{bmatrix} \hat{f}_{11}(k) & \hat{f}_{12}(k) & \cdots & \hat{f}_{1n}(k) \\
            \hat{f}_{21}(k) & \hat{f}_{22}(k) & \cdots & \hat{f}_{2n}(k) \\
	    \vdots & \vdots & \ddots & \vdots \\
            \hat{f}_{n1}(k) & \hat{f}_{n2}(k) & \cdots & \hat{f}_{nn}(k) \\
           \end{bmatrix}.
\]

We refer to the $d_k\times d_k$ block of $X^{(k)}$ corresponding to $\rho_k(g_i) \rho_k^*(g_j) = \rho_k(g_i g_j^{-1})$ as $X^{(k)}_{ij}$.
We now turn our attention to constraints on the variables $\left\{X^{(k)}\right\}_{k=0}^\infty$. It is easy to see that:
\begin{align}
&X^{(k)} \succeq 0, \forall_k\label{cnstrt:psd}\\
&X^{(k)}_{ii} = I_{d_k\times d_k}, \forall_{k,i},\label{cnstrt:identity}\\
&\rank \left[ X^{(k)} \right] = d_k, \forall_{k},\label{cnstrt:rank}\\
&X_{ij}^{(k)} \in \mathrm{Im}(\rho_k), \forall_{k,i,j}.\label{cnstrt:group}
\end{align}

Constraints \eqref{cnstrt:psd}, \eqref{cnstrt:identity} and \eqref{cnstrt:rank} ensure $X^{(k)}$ is of the form
\[
X^{(k)} = \begin{bmatrix} X_1^{(k)} \\ X_2^{(k)} \\ \vdots \\ X_n^{(k)} \end{bmatrix} \begin{bmatrix} X_1^{(k)} \\ X_2^{(k)} \\ \vdots \\ X_n^{(k)} \end{bmatrix}^*,
\]
for some $X^{(k)}_i$ unitary $d_k\times d_k$ matrices.
The constraint \eqref{cnstrt:group} attempts to ensure that $X^{(k)}_i$ is in the image of the representation of $\GGG$.
Notably, none of these constraints ensures that, for different values of $k$, $X^{(k)}_{ij}$ correspond to the same group element $g_i g_j^{-1}$.
Adding such a constraint would yield
\begin{equation}\label{eq:NUG_v2}
\begin{aligned}
& \underset{X^{(k)}}{\operatorname{minimize}} && \sum_{k=0}^\infty \tr\left[ C^{(k)} X^{(k)} \right]\\
& \text{subject to} && X^{(k)} \succeq 0\\
&&& X^{(k)}_{ii} = I_{d_k\times d_k} \\
&&& \rank \left[ X^{(k)} \right] = d_k\\
&&& X_{ij}^{(k)} = \rho_k(g_i g_j^{-1})\text{ for some } g_i g_j^{-1}\in \GGG.
\end{aligned}
\end{equation}

Unfortunately, both the rank constraint and the last constraint in~\eqref{eq:NUG_v2} are, in general, nonconvex. We will relax~\eqref{eq:NUG_v2} by dropping the rank requirement and replacing the last constraint by positivity constraints that couple different $X^{(k)}$'s.
We achieve this by considering the \emph{Dirac delta} funcion on $\GGG$.
Notice that the Dirac delta funcion $\delta(g)$ on the identity $e\in \GGG$ can be expanded as
\[
\begin{aligned}
\delta(g) =& \sum_{k = 0}^{\infty} d_k \tr \left[ \hat{\delta}(k) \rho_k(g) \right]\\
=& \sum_{k = 0}^{\infty} d_k \tr \left[ \left( \int_\GGG \delta(h) \rho_k^*(h) dh \right) \rho_k(g) \right]\\
=& \sum_{k = 0}^{\infty} d_k \tr \left[ \rho_k(g) \right].
\end{aligned}
\]
If we replace $g$ with $g^{-1} \left( g_i g_j^{-1} \right)$, then we get
\[
\begin{aligned}
\delta(g^{-1} g_{ij}) =& \sum_{k = 0}^{\infty} d_k \tr \left[ \rho_k(g^{-1}) \rho_k\left( g_i g_j^{-1} \right) \right]\\
=&  \sum_{k = 0}^{\infty} d_k \tr \left[ \rho_k^*(g) X_{ij}^{(k)} \right].
\end{aligned}
\]
This means that, by the definition of the Dirac delta, we can require that
\begin{align}
&\sum_{k = 0}^{\infty} d_k \tr \left[ \rho_k^*(g) X_{ij}^{(k)} \right] \geq 0 \quad \forall g \in \GGG,\label{eq:Dirac_nonneg}\\
&\int_\GGG \left( \sum_{k = 0}^{\infty} d_k \tr \left[ \rho_k^*(g) X_{ij}^{(k)} \right] \right) dg = 1.\label{eq:Dirac_int1}
\end{align}

This suggests relaxing \eqref{eq:NUG_v2} to
\begin{equation}\label{eq:NUG_v3}
\begin{aligned}
& \underset{X^{(k)}}{\operatorname{minimize}} && \sum_{k=0}^\infty \tr\left[ C^{(k)} X^{(k)} \right]\\
& \text{subject to} && X^{(k)} \succeq 0\\
&&& X^{(k)}_{ii} = I_{d_k\times d_k} \\
&&& \sum_{k = 0}^{\infty} d_k \tr \left[ \rho_k^*(g) X_{ij}^{(k)} \right] \geq 0 \quad \forall g \in \GGG\\
&&& \int_\GGG \left( \sum_{k = 0}^{\infty} d_k \tr \left[ \rho_k^*(g) X_{ij}^{(k)} \right] \right) dg = 1.
\end{aligned}
\end{equation}

For a nontrivial irreducible representation $\rho_k$, we have $\int_\GGG \rho_k(g) dg = 0$.
This means that the integral constraint in~\eqref{eq:NUG_v3} is equivalent to the contraint
\[
 X_{ij}^{(0)} = 1, \forall_{i,j}.
\]
Thus, we focus on the optimization problem
\begin{equation}\label{eq:NUG_SDP}
\begin{aligned}
& \underset{X^{(k)}}{\operatorname{minimize}} && \sum_{k=0}^\infty \tr\left[ C^{(k)} X^{(k)} \right]\\
& \text{subject to} && X^{(k)} \succeq 0\\
&&& X^{(k)}_{ii} = I_{d_k\times d_k} \\
&&& \sum_{k = 0}^{\infty} d_k \tr \left[ \rho_k^*(g) X_{ij}^{(k)} \right] \geq 0 \quad \forall g \in \GGG\\
&&& X_{ij}^{(0)} = 1.
\end{aligned}
\end{equation}

When $\GGG$ is a finite group it has only a finite number of irreducible representations.
This means that~\eqref{eq:NUG_SDP} is a semidefinite program and can be solved, to arbitrary precision, in polynomial time~\cite{LVanderberghe_SBoyd_1996}.
In fact, when $\GGG \cong \ZZ_L$, a suitable change of basis shows that~\eqref{eq:NUG_SDP} is equivalent to the semidefinite programming relaxation proposed in~\cite{Bandeira_MultireferenceAlignment} for the \emph{signal alignment problem}.

Unfortunately, many of the applications of interest involve infinite groups.
This creates two obstacles to solving~\eqref{eq:NUG_SDP}.
One is due to the infinite sum in the objective function and the other due to the infinite number of positivity constraints.
In the next section, we address these two obstacles for the groups $SO(2)$ and $SO(3)$.


\section{Finite truncations for $SO(2)$ and $SO(3)$ via \emph{Fej\'{e}r kernels}}

The objective of this section is to replace~\eqref{eq:NUG_SDP} by an optimization problem depending only in finitely many variables $X^{(k)}$. The objective function in~\eqref{eq:NUG_SDP} is converted from an infinite sum to a finite sum by truncating at degree $t$.
That is, we fix a $t$ and set $C^{(k)} = 0$ for $k > t$. This consists of truncating the Fourier series of $ \sum_{i,j=1}^n f_{ij} \left( g_i g_j^{-1} \right)$. Unfortunately, constraint~\eqref{eq:Dirac_nonneg} given by
\[
 \sum_{k = 0}^{\infty} d_k \tr \left[ \rho_k^*(g) X_{ij}^{(k)} \right] \geq 0 \quad \forall g \in \GGG,
\]
still involves infinitely many variables $X_{ij}^{(k)}$ and consists of infinitely many linear constraints.

We now address this issue for the groups $SO(2)$ and $SO(3)$.

\subsection{Truncation for $SO(2)$}

Since we truncated the objective function at degree $t$, it is then natural to truncate the infinite sum in constraint~\eqref{eq:Dirac_nonneg} also at $t$.
The irreducible representations of $SO(2)$ are $\{ e^{\mathrm{i} k \theta}\}$, and $d_k = 1$ for all $k$.
Let us identify $g \in SO(2)$ with $\theta_g \in [0,2\pi]$.
That straightforward truncation corresponds to approximating the Dirac delta with
\[
\delta(g) \approx \sum_{k = -t}^{t} e^{\mathrm{i} k \theta_g}.
\]
This approximation is known as the \emph{Dirichlet kernel}, which we denote as
\[
D_t(\theta) := \sum_{k = -t}^{t} e^{\mathrm{i} k \theta}.
\]
However, the Dirichlet kernel does not inherit all the desirable properties of the delta function.
In fact, $D_t(\theta)$ is negative for some values of $\theta$.

Instead, we use the \emph{Fej\'{e}r kernel}, which is a non-negative kernel, to approximate the Dirac delta.
The \emph{Fej\'{e}r kernel} is defined as
\[
F_t(\theta) := \frac{1}{t} \sum_{k = 0}^{t-1} D_k = \sum_{k = -t}^t \left( 1 - \frac{|k|}{t}\right) e^{\mathrm{i} k \theta},
\]
which is the first-order \emph{Ces\`{a}ro mean} of the Dirichlet kernel.

This motivates us to replace constraint~\eqref{eq:Dirac_nonneg} with
\[
\sum_{k = -t}^t \left( 1 - \frac{|k|}{t}\right) e^{- \mathrm{i} k \theta} X_{ij}^{(k)} \geq 0 \quad \forall \theta \in [0,2\pi],
\]
where, for $k>0$, $X_{ij}^{(-k)}$ denotes $\left[X_{ij}^{(k)}\right]^\ast$.

This suggests considering
\[
\begin{aligned}
& \underset{X^{(k)}}{\operatorname{minimize}} && \sum_{k=0}^t \tr\left[ C^{(k)} X^{(k)} \right]\\
& \text{subject to} && X^{(k)} \succeq 0\\
&&& X^{(k)}_{ii} = I_{d_k\times d_k} \\
&&& \sum_{k = -t}^t \left( 1 - \frac{|k|}{t}\right) e^{- \mathrm{i} k \theta} X_{ij}^{(k)} \geq 0 \quad \forall \theta \in [0,2\pi]\\
&&& X_{ij}^{(0)} = 1,
\end{aligned}
\]
which only depends on the variables $X_{ij}^{(k)}$ for $k=0,\dots,k$.

Unfortunately, the condition that the trigonometric polynomial
\(
 \sum_{k = -t}^t \left( 1 - \frac{|k|}{t}\right) e^{- \mathrm{i} k \theta} X_{ij}^{(k)} 
\)
is always non-negative, still involves an infinite number of linear inequalities.
Interestingly, due to the Fej\'er-Riesz factorization theorem (see~\cite{Dumitrescu_2007}), this condition can be replaced by an equivalent condition involving a positive semidefinite matrix --- it turns out that every nonnegative trigonometric polynomial is a square, meaning that the so called \emph{sum-of-squares relaxation}~\cite{Parrilo00structuredsemidefinite,RekhaThomasbook_2013} is exact.
However, while such a formulation would still be an SDP and thus solvable, up to arbitrary precision, in polynomial time, it would involve a positive semidefinite variable for every pair $(i,j)$, rendering it computationally challenging.
For this reason we relax the non-negativity constraint by asking that
\(
 \sum_{k = -t}^t \left( 1 - \frac{|k|}{t}\right) e^{- \mathrm{i} k \theta} X_{ij}^{(k)} 
\)
is non-negative in a finite set $\Omega_t\in SO(2)$.
This yields the following optimization problem:

%

\begin{equation}\label{eq:NUG_SDP_SO2}
\begin{aligned}
& \underset{X^{(k)}}{\operatorname{minimize}} && \sum_{k=0}^t \tr\left[ C^{(k)} X^{(k)} \right]\\
& \text{subject to} && X^{(k)} \succeq 0\\
&&& X^{(k)}_{ii} = I_{d_k\times d_k} \\
&&& \sum_{k = -t}^t \left( 1 - \frac{|k|}{t}\right) e^{- \mathrm{i} k \theta} X_{ij}^{(k)} \geq 0 \quad \forall \theta \in \Omega_t\\
&&& X_{ij}^{(0)} = 1.
\end{aligned}
\end{equation}

\subsection{Truncation for $SO(3)$}

The irreducible representations of $SO(3)$ are the \emph{Wigner-D matrices} $\{ W^{(k)}(\alpha,\beta,\gamma) \}$, and $d_k = 2 k + 1$.
Let us associate $g \in SO(3)$ with \emph{Euler (Z-Y-Z) angle} $(\alpha,\beta,\gamma) \in [0,2\pi] \times [0,\pi] \times [0,2\pi]$.
A straightforward truncation yields the approximation
\[
\delta(g) \approx \sum_{k = 0}^{t} (2 k + 1) \tr \left[ W^{(k)} (\alpha,\beta,\gamma) \right].
\]
Observe that the operator $\tr$ is invariant under conjugation.
Then $W^{(k)}$ can be decomposed as
\[
W^{(k)}(\alpha,\beta,\gamma) = R \Lambda^{(k)}(\theta) R^*
\]
such that
\[
\Lambda^{(k)}(\theta) = \begin{bmatrix} e^{-\mathrm{i} k \theta} & & & & \\ & \ddots & & & \\ & & 1 & & \\ & & & \ddots & \\ & & & & e^{\mathrm{i} k \theta} \end{bmatrix}.
\]
It follows that
\[
\tr \left[ W^{(k)} (\alpha,\beta,\gamma) \right] = \tr \left[ \Lambda^{(k)}(\theta) \right] = \sum_{l = -k}^k e^{\mathrm{i} m \theta} = D_k(\theta).
\]
The relationship between $\theta$ and $\alpha,\beta,\gamma$ is
\[
\theta = 2 \arccos \left[ \cos \left( \frac{\beta}{2} \right) \cos \left( \frac{\alpha + \gamma}{2} \right) \right].
\]
This relationship can obtained by directly evaluating $\tr \left[ W^{(1)} (\alpha,\beta,\gamma) \right]$ using the \emph{Wigner-d matrix} $w^{(1)}$:
\[
\begin{aligned}
\tr \left[ W^{(1)} (\alpha,\beta,\gamma) \right] =& \sum_{m = -1}^1 W_{m,m}^{(1)} (\alpha,\beta,\gamma)\\
=& \sum_{m = -1}^1 e^{- \mathrm{i} m (\alpha + \gamma)} w_{m,m}^{(1)}(\beta)\\
=& \cos ( \beta ) \left( 1 + \cos ( \alpha + \gamma ) \right) + \cos ( \alpha + \gamma ).
\end{aligned}
\]
This straightfoward truncation at $t$ yields
\[
\delta(g) \approx \sum_{k = 0}^{t} (2 k + 1) D_k(\theta),
\]
which, again, inherits the undesirable property that this approximation can be negative for some $\theta$.
Recall that we circumvented this property in the $1$-dimension case by taking the first-order Ces\`{a}ro mean of the Dirichlet kernel.
In the $2$-dimension case, we will need the second-order Ces\`{a}ro mean.
Notice that
\[
D_k(\theta) = \frac{\sin \left[ (2 k + 1) \frac{\theta}{2} \right]}{\sin \left( \frac{\theta}{2} \right)}.
\]
Fej\'{e}r proved that~\cite{Askey_1975}
\[
\sum_{k=0}^t \frac{(3)_{t-k}}{(t-k)!} \left( k + \frac{1}{2} \right) \sin \left[ (2 k + 1) \frac{\theta}{2} \right] \geq 0, \quad 0 \leq \theta \leq \pi
\]
where $\frac{(3)_{t-k}}{(t-k)!} = \frac{1}{2} (t - k + 2)(t - k + 1)$.
It follows that
\[
\begin{aligned}
&\sum_{k=0}^t \frac{(3)_{t-k}}{(t-k)!} \left( k + \frac{1}{2} \right) D_k(\theta)\\
=& \sum_{k=0}^t \frac{(3)_{t-k}}{(t-k)!} \left( k + \frac{1}{2} \right) \frac{\sin \left[ (2 k + 1) \frac{\theta}{2} \right]}{\sin \left( \frac{\theta}{2} \right)} \geq 0, \quad -\pi \leq 0 \leq \pi.
\end{aligned}
\]
Let us define
\[
F_t(g) = F_t(\alpha,\beta,\gamma) := \sum_{k=0}^t \frac{(3)_{t-k}}{(t-k)!} \left( k + \frac{1}{2} \right) \frac{\sin \left[ (2 k + 1) \frac{\theta_g}{2} \right]}{\sin \left( \frac{\theta_g}{2} \right)}
\]
where $\theta_g = 2 \arccos \left[ \cos \left( \frac{\beta}{2} \right) \cos \left( \frac{\alpha + \gamma}{2} \right) \right]$.
Also, let us define
\[
B_t := \int_{SO(3)} F_t(g) dg,
\]
where $dg$ is the Haar measure.

We replace constraint \eqref{eq:Dirac_nonneg} with
\[
\frac{1}{B_t} F_t(\alpha,\beta,\gamma) \geq 0 \quad \forall (\alpha,\beta,\gamma) \in [0,2\pi] \times [0,\pi] \times [0,2\pi].
\]

Secondly, we discretize the group $SO(3)$ to obtain a finite number of constraints. We consider a suitable finite subset $\Omega_t \subset SO(3)$. We can then relax the non-negativity constraint yielding the following semidefinite program\footnote{Similarly to $SO(2)$, it is possible that the non-negativity constraint may be replaced by and SDP or sums-of-squares constraint.}.
%
%
\begin{equation}\label{eq:NUG_SDP_SO3}
\begin{aligned}
& \underset{X^{(k)}}{\operatorname{minimize}} && \sum_{k=0}^t \tr\left[ C^{(k)} X^{(k)} \right]\\
& \text{subject to} && X^{(k)} \succeq 0\\
&&& X^{(k)}_{ii} = I_{d_k\times d_k} \\
&&& \frac{1}{A_t} \sum_{k = 0}^{t} \frac{(3)_{t-k}}{(t-k)!} \left( k + \frac{1}{2} \right) \tr \left[ \left( W^{(k)} (\alpha,\beta,\gamma) \right)^* X_{ij}^{(k)}\right] \geq 0 \quad \forall (\alpha,\beta,\gamma)\in \Omega_t\\
&&& X_{ij}^{(0)} = 1.
\end{aligned}
\end{equation}

\section{Applications}

In this section, we estimate the solution to registration in $1$-dimension using~\eqref{eq:NUG_SDP_SO2}, and the solutions to registration in $2$-dimensions and orientation estimation in cryo-EM using~\eqref{eq:NUG_SDP_SO3}.
For each problem, the only parameters we need to determine are the coefficient matrices $C^{(k)}$.
Since $C^{(k)} = \left( \hat{f}_{ij}(k) \right)_{i,j=1}^n$, then it suffices to calculate the Fourier coefficients $\hat{f}_{ij}(k)$ for the respective problems.

\subsection{Registration in $1$-dimension}

Recall that $\XXX$ is the space of bandlimited functions up to degree $t$ on $S^1$.
That is, for $x \in \XXX$, we can express
\[
x(\omega) = \sum_{l = -t}^t \alpha_l e^{\mathrm{i} l \omega}.
\]
Again, the irreducible representations of $SO(2)$ are $\{ e^{\mathrm{i} k \theta}\}$, and $d_k = 1$ for all $k$.
Let us identify $g \in SO(2)$ with $\theta_g \in [0,2\pi]$, then
\[
g \cdot x(\omega) = \sum_{l = -t}^t \alpha_l e^{\mathrm{i} l \omega} e^{-\mathrm{i} l \theta} = \sum_{l = -t}^t \alpha_l e^{\mathrm{i} l (\omega - \theta)}.
\]
Let $\PPP$ sample the underlying signal $x$ at $L = 2 t + 1$ distinct points.
This way, we can determine all the $\alpha_l$'s associated with $x$.

Since $y_i = \PPP (g_i \cdot x) + \epsilon_i$, then we can approximate $y_i$ with the expansion
\[
\mathcal{Q}(y_i)(\omega) \approx \sum_{l=-t}^t \alpha_l^{(i)} e^{\mathrm{i} l \omega}.
\]

Let us identify $g_i g_j^{-1}$ with $\theta_{ij} \in [0,2\pi]$.
Then, we can express $f_{ij}$ in terms of $\alpha_l^{(i)}$, $\alpha_l^{(j)}$ and $\theta_{ij}$:
\[
\begin{aligned}
f_{ij} \left(g_i g_j^{-1} \right) =& \left\| y_i - g_{ij} \cdot y_j \right\|_2^2\\
=& \int_{S^1} \left| \sum_{l = -t}^t \left( \alpha_l^{(i)} - \alpha_l^{(j)} e^{-\mathrm{i} l \theta_{ij}} \right) e^{\mathrm{i} l \omega} \right|^2d\omega\\
=& \sum_{l = -t}^t \left| \alpha_l^{(i)} - \alpha_l^{(j)} e^{-\mathrm{i} l \theta_{ij}} \right|^2.
\end{aligned}
\]

The Fourier coefficients of $f_{ij}$ are
\[
\begin{aligned}
\hat{f}_{ij} (k) =& \int_{SO(2)} \sum_{l = -t}^t \left| \alpha_l^{(i)} - \alpha_l^{(j)} e^{-\mathrm{i} l \theta_{ij}} \right|^2 e^{\mathrm{i} k \theta_{ij}} dg\\
=& \int_{0}^{2 \pi} \sum_{l = -t}^t \left| \alpha_l^{(i)} - \alpha_l^{(j)} e^{-\mathrm{i} l \theta_{ij}} \right|^2 e^{\mathrm{i} k \theta_{ij}} d\theta_{ij}\\
=& \int_{0}^{2 \pi} \sum_{l = -t}^t \left( | \alpha_l^{(i)} |^2 e^{\mathrm{i} k \theta_{ij}} + | \alpha_l^{(j)} |^2 e^{\mathrm{i} k \theta_{ij}} - \overline{\alpha_l^{(i)}} \alpha_l^{(j)} e^{\mathrm{i} (k - l) \theta_{ij}} - \alpha_l^{(i)} \overline{\alpha_l^{(j)}} e^{\mathrm{i} (k + l) \theta_{ij}} \right) d\theta_{ij}\\
=& 2 \pi \begin{cases}
\sum_{l = -t}^t \left( | \alpha_l^{(i)} |^2 + | \alpha_l^{(j)} |^2 \right) &, \quad k = 0 \\
\\
- \overline{\alpha_k^{(i)}} \alpha_k^{(j)} - \alpha_{-k}^{(i)} \overline{\alpha_{-k}^{(j)}} &, \quad k \neq 0
\end{cases}
\end{aligned}
\]
Note that we re-indexed the coefficients $\hat{f}_{ij}(k) \leftarrow \hat{f}_{ij}(k - (t+1))$.

\subsection{Registration in $2$-dimension}

Recall that $\XXX$ is the space of bandlimited functions up to degree $t$ on $S^2$.
That is, for $x \in \XXX$, we can express
\[
x(\omega) = \sum_{l = 0}^t \sum_{m = -l}^l \alpha_{lm} Y_{lm} (\omega),
\]
where $\{ Y_{lm} \}$ are the spherical harmonics.
Again, the irreducible representations of $SO(3)$ on are the Wigner-D matrices $\{ W^{(k)}(\alpha,\beta,\gamma) \}$, and $d_k = 2 k + 1$.
Let us associate $g \in SO(3)$ with Euler (Z-Y-Z) angle $(\alpha,\beta,\gamma) \in [0,2\pi] \times [0,\pi] \times [0,2\pi]$, then
\[
g \cdot x(\omega) = \sum_{l = 0}^t \sum_{m,m' = -l}^l W_{m,m'}^{(l)}(\alpha,\beta,\gamma) \alpha_{lm'} Y_{lm}(\omega).
\]
Let $\PPP$ sample the underlying signal $x$ at $L = (t + 1)^2$ points.
This way, we can determine all the $\alpha_{lm}$'s associated with $x$.

Since $y_i = \PPP (g_i \cdot x) + \epsilon_i$, then we can approximate $y_i$ with the expansion
\[
\mathcal{Q}(y_i)(\omega) \approx \sum_{l=0}^t \sum_{m=-l}^l \alpha_{lm}^{(i)} Y_{lm}(\omega).
\]

Let us identify $g_i g_j^{-1} \in SO(3)$ with Euler (Z-Y-Z) angle $(\alpha_{ij},\beta_{ij},\gamma_{ij}) \in [0,2\pi] \times [0,\pi] \times [0,2\pi]$.
Then, we can express $f_{ij}$ in terms of $\alpha_{lm}^{(i)}$, $\alpha_{lm}^{(j)}$ and $(\alpha_{ij},\beta_{ij},\gamma_{ij})$:
\[
\begin{aligned}
f_{ij} \left( g_i g_j^{-1} \right) =& \left\| y_i - g_i g_j^{-1} \cdot y_j \right\|_2^2\\
=& \sum_{l = 0}^t \sum_{m = -l}^l \left( | \alpha_{lm}^{(i)} |^2 + | \alpha_{lm}^{(j)} |^2 \right) - 2 \sum_{l = 0}^t \sum_{m,m' = -l}^l \alpha_{lm}^{(i)} \overline{W_{m,m'}^{(l)}} (\alpha_{ij},\beta_{ij},\gamma_{ij}) \overline{\alpha_{lm'}^{(j)}}.
\end{aligned}
\]

The Fourier coefficients are given by
\[
\begin{aligned}
\hat{f}_{ij} (k) =& \int_{SO(3)} \left[ \sum_{l = 0}^t \sum_{m = -l}^l \left( | \alpha_{lm}^{(i)} |^2 + | \alpha_{lm}^{(j)} |^2 \right) - 2 \sum_{l = 0}^t \sum_{m,m' = -l}^l \alpha_{lm}^{(i)} \overline{W_{m,m'}^{(l)}} (\alpha,\beta,\gamma) \overline{\alpha_{lm'}^{(j)}} \right] W^{(k)} (\alpha,\beta,\gamma) dg\\
=& \frac{8 \pi^2}{2 k + 1} \begin{cases}
\sum_{l = 0}^t \sum_{m = -l}^l \left( | \alpha_{lm}^{(i)} |^2 + | \alpha_{lm}^{(j)} |^2 \right) - 2 \alpha_{00}^{(i)} \overline{\alpha_{00}^{(j)}} &, \quad k = 0 \\
\\
\left( - 2 \alpha_{km}^{(i)} \overline{\alpha_{km'}^{(j)}}\right)_{m,m' = -k}^k &, \quad k \neq 0
\end{cases}
\end{aligned}
\]
Here, we used the orthogonality relationship
\[
\int_{SO(3)} \overline{W_{l,m}^{(k)}} (\alpha,\beta,\gamma) W_{l',m'}^{(k')} (\alpha,\beta,\gamma) dg = \frac{8 \pi^2}{2 k + 1} \delta_{k,k'} \delta_{l,l'} \delta_{m,m'}.
\]

\subsection{Orientation estimation in cryo-EM}

We refer to \cite{Zhao_FastSteerablePCA} to expand the objective function; projection $\hat{I}_i$ can be expanded via \emph{Fourier-Bessel series} as
\[
\hat{I}_i (r,\theta) = \sum_{k = -\infty}^{\infty} \sum_{q = 1}^{\infty} \alpha_{kq}^{(i)} \psi_{kq}^{(c)} (r,\theta),
\]
where
\[
\psi_{kq}^{(c)} (r,\theta) = \begin{cases} N_{kq} J_k \left( R_{kq} \frac{r}{c} \right) e^{i k \theta} &, \quad r \leq c, \\ 0 &, \quad r > c. \end{cases}
\]
The parameters above are defined as follows:
\begin{itemize}
\item $c$ is the radius of the disc containing the support of $I_i$ (recall $\phi \in X$ has compact support).
\item $J_k$ is the \emph{Bessel function} of integer order $k$,
\item $R_{kq}$ is the $q^{th}$ root of $J_k$,
\item $N_{kq} = \frac{1}{c \sqrt{\pi} | J_{k+1}(R_{kq}) |}$ is a normalization factor.
\end{itemize}
We can approximate each \emph{Fourier-Bessel} expansion by truncating.
I.e.,
\[
\hat{I}_i (r,\theta) \approx \sum_{k = -k_{\text{max}}}^{k_{\text{max}}} \sum_{q = 1}^{p_k} \alpha_{kq}^{(i)} \psi_{kq}^{(c)} (r,\theta).
\]
See \cite{Zhao_FastSteerablePCA} for a discussion on $k_{\text{max}}$ and $p_k$.
For the purpose of this section, let us assume we have $\{ \alpha_{kq}^{(i)} : -k_{\text{max}} \leq k \leq k_{\text{max}}, 1 \leq q \leq p_k\}$ for each $\hat{I}_i$.
(These can be computed from data.)

We shall determine the relationship between $\hat{I}_i (r,\theta_i)$ and $\hat{I}_j (r,\theta_j)$, and the lines of intersection between $g_i^{-1} \cdot \hat{I}_i$ and $g_j^{-1} \cdot \hat{I}_j$ embedded in $\RR^3$.
Recall from \eqref{eq:cryoEM_cline_i} and \eqref{eq:cryoEM_cline_j} that the directions of the lines of intersection between $g_i^{-1} \cdot \hat{I}_i$ and $g_j^{-1} \cdot \hat{I}_j$ are given, respectively, by unit vectors
\[
\begin{aligned}
c_{ij} \left( g_i g_j^{-1} \right) =& \frac{ \vec{e}_3 \times g_i g_j^{-1} \cdot \vec{e}_3 }{\| \vec{e}_3 \times g_i g_j^{-1} \cdot \vec{e}_3 \|},\\
c_{ji} \left( g_i g_j^{-1} \right) =& \frac{\vec{e}_3 \times \left( g_i g_j^{-1} \right)^{-1} \cdot \vec{e}_3}{\| \vec{e}_3 \times \left( g_i g_j^{-1} \right)^{-1} \cdot \vec{e}_3 \|}.
\end{aligned}
\]
Let us associate $g_i g_j^{-1} \in SO(3)$ with Euler (Z-Y-Z) angle $(\alpha_{ij},\beta_{ij},\gamma_{ij}) \in [0,2\pi] \times [0,\pi] \times [0,2\pi]$.
Then
\[
\begin{aligned}
\vec{e}_3 \times g_i g_j^{-1} \cdot \vec{e}_3 =& \begin{bmatrix} -\sin \alpha_{ij} \sin \beta_{ij} \\ - \cos \alpha_{ij} \sin \beta_{ij} \\ 0 \end{bmatrix},\\
\vec{e}_3 \times \left( g_i g_j^{-1} \right)^{-1} \cdot \vec{e}_3 =& \begin{bmatrix} \sin \gamma_{ij} \sin \beta_{ij} \\ \cos \gamma_{ij} \sin \beta_{ij} \\ 0 \end{bmatrix}.
\end{aligned}
\]
The directions of the lines of intersection in $\hat{I}_i$ and $\hat{I}_j$ under $g_i g_j^{-1}$ are in the directions, respectively,
\[
\begin{aligned}
\theta_i =& \arctan \left( \frac{\cos \alpha_{ij}}{\sin \alpha_{ij}} \right) = \frac{\pi}{2} - \alpha_{ij},\\
\theta_j =& \arctan \left( \frac{\cos \gamma_{ij}}{\sin \gamma_{ij}} \right) = \frac{\pi}{2} - \gamma_{ij}.
\end{aligned}
\]

We express the $f_{ij}$'s in terms of $\alpha_{kq}^{(i)}$, $\alpha_{kq}^{(j)}$, and $\theta_i$ and $\theta_j$:
\[
\begin{aligned}
f_{ij}(\theta_i,\theta_j) :=& f_{ij} \left( g_i g_j^{-1} \right)\\
=& \left\| \sum_{k = -k_{\text{max}}}^{k_{\text{max}}} \sum_{q = 1}^{p_k} \left( \alpha_{kq}^{(i)} \psi_{kq}^{(c)} (r,\theta_i) - \alpha_{kq}^{(j)} \psi_{kq}^{(c)} (r,\theta_j) \right) \right\|_{L^2}^2 \\
=& \sum_{k,k',q,q'} c N_{kq} N_{k'q'} \left( \alpha_{kq}^{(i)} e^{\mathrm{i} k \theta_i} - \alpha_{kq}^{(j)} e^{\mathrm{i} k \theta_j} \right) \left( \alpha_{k'q'}^{(i)} e^{\mathrm{i} k' \theta_i} - \alpha_{k'q'}^{(j)} e^{\mathrm{i} k' \theta_j} \right)^* \int_0^1 J_k(R_{kq} r) J_k'(R_{k'q'} r) dr.
\end{aligned}
\]
For each $k,k',q,q'$, we approximate the integral
\[
\int_0^1 J_k(R_{kq} r) J_k'(R_{k'q'} r) dr
\]
with a Gaussian quadrature using the roots of $P_{2s}(\sqrt{x})$ and weights $\frac{1}{\sqrt{x}}$.
Here, $P_{2s}$ is the Legendre polynomial and we specify a suitable $s$.

Using the approximation above, we have
\[
\begin{aligned}
f_{ij}(\theta_i,\theta_j) \approx \sum_{k,k',q,q'} &b_{k,k',q,q'} \Biggl( \alpha_{kq}^{(i)} \overline{\alpha_{k'q'}^{(i)}} e^{\mathrm{i} (k - k') \theta_i} + \alpha_{kq}^{(j)} \overline{\alpha_{k'q'}^{(j)}} e^{\mathrm{i} (k - k') \theta_j} \\
& - \alpha_{kq}^{(i)} \overline{\alpha_{k'q'}^{(j)}} e^{\mathrm{i} (k \theta_i - k' \theta_j)} - \alpha_{kq}^{(j)} \overline{\alpha_{k'q'}^{(i)}} e^{\mathrm{i} (k \theta_j - k' \theta_i)} \Biggr),
\end{aligned}
\]
where
\[
b_{k,k',q,q'} = c N_{kq} N_{k'q'} \sum_{x_i} \frac{J_k(R_{kq} x_i) J_k'(R_{k'q'} x_i)}{\sqrt{x_i}}
\]
and $x_i$'s are the roots of $P_{2s}(\sqrt{x})$.

In terms of the Euler (Z-Y-Z) angles,
\[
\begin{aligned}
f_{ij}(\alpha_{ij},\gamma_{ij}) :=& f_{ij}(\theta_i,\theta_j)\\
\approx& \sum_{k,k',q,q'} b_{k,k',q,q'} e^{\mathrm{i} \frac{\pi}{2} (k - k')} \Biggl( \alpha_{kq}^{(i)} \overline{\alpha_{k'q'}^{(i)}} e^{- \mathrm{i} (k - k') \alpha_{ij}} + \alpha_{kq}^{(j)} \overline{\alpha_{k'q'}^{(j)}} e^{- \mathrm{i} (k - k') \gamma_{ij}} \\
& - \alpha_{kq}^{(i)} \overline{\alpha_{k'q'}^{(j)}} e^{\mathrm{i} (-k \gamma_{ij} + k' \alpha_{ij})} - \alpha_{kq}^{(j)} \overline{\alpha_{k'q'}^{(i)}} e^{\mathrm{i} (-k \alpha_{ij} + k' \gamma_{ij})} \Biggr).
\end{aligned}
\]

The Fourier coefficients are by
\[
\begin{aligned}
\hat{f}_{ij}(k) =& \int_{SO(3)} f_{ij}(\alpha,\gamma) W^{(k)} (\alpha,\beta,\gamma) dg\\
=& \int_0^{2 \pi} \int_0^{2 \pi} f_{ij}(\alpha,\gamma) \left( \int_0^{\pi} W^{(k)} (\alpha,\beta,\gamma) \sin \beta d\beta \right) d\alpha d\gamma.
\end{aligned}
\]
The $(m,m')^{th}$ entry of $\hat{f}_{ij}(k)$ is approximated by
\[
\begin{aligned}
\left( \hat{f}_{ij}(k) \right)_{m,m'} &\approx \int_0^{2 \pi} \int_0^{2 \pi} f_{ij}(\alpha,\gamma) \left( \int_0^{\pi} e^{-\mathrm{i} m \alpha}  w_{m,m'}^{(k)}(\beta) e^{-\mathrm{i} m' \gamma} \sin \beta d\beta \right) d\alpha d\gamma\\
&= 2 w_{m,m'}^{(k)}(\pi/2) \int_0^{2 \pi} \int_0^{2 \pi} f_{ij}(\alpha,\gamma) e^{-\mathrm{i} m \alpha} e^{-\mathrm{i} m' \gamma} d\alpha d\gamma\\
&\begin{aligned}
= 8 \pi^2 w_{m,m'}^{(k)}(\pi/2) \sum_{k_1,k_2,q_1,q_2} b_{k_1,k_2,q_1,q_2} \Biggl( &\alpha_{k_1 q_1}^{(i)} \overline{\alpha_{k_2 q_2}^{(i)}} \delta_{0,m'} \delta_{0,k_1-k_2+m}\\
+& \alpha_{k_1 q_1}^{(j)} \overline{\alpha_{k_2 q_2}^{(j)}} \delta_{0,m} \delta_{0,k_1-k_2+m'}\\
-& \alpha_{k_1 q_1}^{(i)} \overline{\alpha_{k_2 q_2 }^{(j)}} \delta_{k_1,-m'} \delta_{k_2,-m}\\
-& \alpha_{k_1 q_1}^{(j)} \overline{\alpha_{k_2 q_2}^{(i)}} \delta_{k_1,-m} \delta_{k_2,-m'} \Biggr).
\end{aligned}
\end{aligned}
\]
Here, $w^{(k)}$ is the Wigner-d matrix, $\delta$ is the \emph{Kronecker delta} and $b_{k_1,k_2,q_1,q_2}$ absorbed $e^{\mathrm{i} \frac{\pi}{2} (k_1 - k_2)}$.

\section{Conclusion}\label{sec:discussion}

This short manuscript is a preliminary version of a future publication with the same title and same authors.
In particular, we defer an extensive numerical study of this approach to the future publication.
It is worth mentioning that preliminary numerical simulations suggest that this approach has a tendency to be tight, meaning that the solution to the relaxation often coincides with the solution to the original NUG problem under certain nonadversarial noise models. This tendency for certain semidefinite relaxations was conjectured in~\cite{Bandeira_TigtnessMLE_OpenProblem} and established in~\cite{Bandeira_rankrecoveryangsynch} for a particularly simple instance of angular synchronization.

In many applications, such as the structure from motion problem in Computer Vision~\cite{matinec2007rotation,hartley2013rotation}, the group $\GGG$ is non-compact. However, such groups can often be compactified by mapping a subset of $\GGG$ to a compact group. One example is to treat the group of Euclidean motions in $d$ dimensions by mapping a bounded subset to $SO(d+1)$ (see~\cite{ASinger_2011_angsync} for a description of the case $d=1$).

\subsection*{Acknowledgements}

The authors would like to thank Moses Charikar and Peter Sarnak for many insightful discussions on the topic of this paper.


 \bibliographystyle{plain}
 \bibliography{NUG_bib}

\begin{thebibliography}{10}

\bibitem{Abbe_Z2Synch}
E.~Abbe, A.~S. Bandeira, A.~Bracher, and A.~Singer.
\newblock Decoding binary node labels from censored edge measurements: Phase
  transition and efficient recovery.
\newblock {\em Network Science and Engineering, IEEE Transactions on},
  1(1):10--22, Jan 2014.

\bibitem{Abbe_SBMExact}
E.~Abbe, A.~S. Bandeira, and G.~Hall.
\newblock Exact recovery in the stochastic block model.
\newblock {\em Available online at arXiv:1405.3267 [cs.SI]}, 2014.

\bibitem{NAlon_ANaor_2006}
N.~Alon and A.~Naor.
\newblock Approximating the cut-norm via {G}rothendieck's inequality.
\newblock In {\em Proc. of the 36 th ACM STOC}, pages 72--80. ACM Press, 2004.

\bibitem{Askey_1975}
R.~Askey.
\newblock {\em Orthogonal Polynomials and Special Functions}.
\newblock SIAM, 4 edition, 1975.

\bibitem{Bandeira_Laplacian}
A.~S. Bandeira.
\newblock Random {L}aplacian matrices and convex relaxations.
\newblock {\em Available online at arXiv:1504.03987 [math.PR]}, 2015.

\bibitem{Bandeira_rankrecoveryangsynch}
A.~S. Bandeira, N.~Boumal, and A.~Singer.
\newblock Tightness of the maximum likelihood semidefinite relaxation for
  angular synchronization.
\newblock {\em Available online at arXiv:1411.3272 [math.OC]}, 2014.

\bibitem{Bandeira_MultireferenceAlignment}
A.~S. Bandeira, M.~Charikar, A.~Singer, and A.~Zhu.
\newblock Multireference alignment using semidefinite programming.
\newblock {\em 5th Innovations in Theoretical Computer Science (ITCS 2014)},
  2014.

\bibitem{Bandeira_LittleGrothendieckOd}
A.~S. Bandeira, C.~Kennedy, and A.~Singer.
\newblock Approximating the little {G}rothendieck problem over the orthogonal
  and unitary groups.
\newblock {\em Available online at arXiv:1308.5207 [cs.DS]}, 2013.

\bibitem{Bandeira_TigtnessMLE_OpenProblem}
A.~S. Bandeira, Y.~Khoo, and A.~Singer.
\newblock Open problem: Tightness of maximum likelihood semidefinite
  relaxations.
\newblock In {\em Proceedings of the 27th Conference on Learning Theory},
  volume~35 of {\em JMLR W\&CP}, pages 1265--1267, 2014.

\bibitem{Bandeira_Singer_Spielman_OdCheeger}
A.~S. Bandeira, A.~Singer, and D.~A. Spielman.
\newblock A {C}heeger inequality for the graph connection {L}aplacian.
\newblock {\em SIAM J. Matrix Anal. Appl.}, 34(4):1611--1630, 2013.

\bibitem{Charikar_Makarychev_UG}
M.~Charikar, K.~Makarychev, and Y.~Makarychev.
\newblock Near-optimal algorithms for unique games.
\newblock {\em Proceedings of the 38th ACM Symposium on Theory of Computing},
  2006.

\bibitem{Chen_Huang_Guibas_Graphics}
Y.~Chen, Q.-X. Huang, and L.~Guibas.
\newblock Near-optimal joint object matching via convex relaxation.
\newblock {\em Available Online: arXiv:1402.1473 [cs.LG]}.

\bibitem{Cohen_ModelBias}
J.~Cohen.
\newblock Is high-tech view of {HIV} too good to be true?
\newblock {\em Science}, 341(6145):443--444, 2013.

\bibitem{Cucuringu_Z2Synch}
M.~Cucuringu.
\newblock Synchronization over {Z}2 and community detection in signed multiplex
  networks with constraints.
\newblock {\em Journal of Complex Networks}, 2015.

\bibitem{Dumitrescu_2007}
B.~Dumitrescu.
\newblock {\em Positive Trigonometric Polynomials and Signal Processing
  Applications}.
\newblock Springer, 2007.

\bibitem{RekhaThomasbook_2013}
P.~A.~Parrilo G.~Blekherman and R.~R. Thomas.
\newblock {\em Semidefinite Optimization and Convex Algebraic Geometry}.
\newblock MOS-SIAM Series on Optimization, 2013.

\bibitem{MXGoemans_DPWilliamson_1995}
M.~X. Goemans and D.~P. Williamson.
\newblock Improved apprximation algorithms for maximum cut and satisfiability
  problems using semidefine programming.
\newblock {\em Journal of the Association for Computing Machinery},
  42:1115--1145, 1995.

\bibitem{hartley2013rotation}
R.~Hartley, J.~Trumpf, Y.~Dai, and H.~Li.
\newblock Rotation averaging.
\newblock {\em International Journal of Computer Vision}, 103(3):267--305,
  2013.

\bibitem{Huang_Guibas_Graphics}
Q.-X. Huang and L.~Guibas.
\newblock Consistent shape maps via semidefinite programming.
\newblock {\em Computer Graphics Forum}, 32(5):177--186, 2013.

\bibitem{SKhot_2002}
S.~Khot.
\newblock On the power of unique 2-prover 1-round games.
\newblock {\em Thiry-fourth annual ACM symposium on Theory of computing}, 2002.

\bibitem{SKhot_2010}
S.~Khot.
\newblock On the unique games conjecture (invited survey).
\newblock In {\em Proceedings of the 2010 IEEE 25th Annual Conference on
  Computational Complexity}, CCC '10, pages 99--121, Washington, DC, USA, 2010.
  IEEE Computer Society.

\bibitem{matinec2007rotation}
D.~Martinec and T.~Pajdla.
\newblock Robust rotation and translation estimation in multiview
  reconstruction.
\newblock In {\em IEEE Conference on Computer Vision and Pattern Recognition,
  2007. CVPR '07}, pages 1--8, June 2007.

\bibitem{Naor_etal_NCGI}
A.~Naor, O.~Regev, and T.~Vidick.
\newblock Efficient rounding for the noncommutative {G}rothendieck inequality.
\newblock In {\em Proceedings of the 45th annual ACM symposium on Symposium on
  theory of computing}, STOC '13, pages 71--80, New York, NY, USA, 2013. ACM.

\bibitem{book:Natterer}
F.~Natterer.
\newblock {\em The Mathematics of Computerized Tomography}.
\newblock Classics in Applied Mathematics, SIAM, 2001.

\bibitem{Nesterov_quadprogram1}
Y.~Nesterov.
\newblock Semidefinite relaxation and nonconvex quadratic optimization.
\newblock {\em Optimization Methods and Software}, 9(1-3):141--160, 1998.

\bibitem{Parrilo00structuredsemidefinite}
P.~A. Parrilo.
\newblock Structured semidefinite programs and semialgebraic geometry methods
  in robustness and optimization.
\newblock {\em Thesis, California Institute of Technology}, 2000.

\bibitem{ASinger_2011_angsync}
A.~Singer.
\newblock Angular synchronization by eigenvectors and semidefinite programming.
\newblock {\em Appl. Comput. Harmon. Anal.}, 30(1):20 -- 36, 2011.

\bibitem{ASinger_YShkolnisky_commonlines}
A.~Singer and Y.~Shkolnisky.
\newblock Three-dimensional structure determination from common lines in
  {C}ryo-{EM} by eigenvectors and semidefinite programming.
\newblock {\em SIAM J. Imaging Sciences}, 4(2):543--572, 2011.

\bibitem{ASinger_ZZhao_YShkolnisky_RHadani_cryo}
A.~Singer, Z.~Zhao, , Y.~Shkolnisky, and R.~Hadani.
\newblock Viewing angle classification of cryo-{E}lectron {M}icroscopy images
  using eigenvectors.
\newblock {\em SIAM Journal on Imaging Sciences}, 4:723--759, 2011.

\bibitem{Vainshtein_Goncharov}
B.~K. Vainshtein and A.~B. Goncharov.
\newblock Determination of the spatial orientation of arbitrarily arranged
  identical particles of unknown structure from their projections.
\newblock 1986.

\bibitem{vanheel_angularreconstitution}
M.~Van~Heel.
\newblock Angular reconstitution: a posteriori assignment of projection
  directions for 3d reconstruction.
\newblock {\em Ultramicroscopy}, 21:111--123, 1987.

\bibitem{LVanderberghe_SBoyd_1996}
L.~Vanderberghe and S.~Boyd.
\newblock Semidefinite programming.
\newblock {\em SIAM Review}, 38:49--95, 1996.

\bibitem{Zhao_FastSteerablePCA}
Z.~Zhao, Y.~Shkolnisky, and A.~Singer.
\newblock Fast steerable principal component analysis.
\newblock {\em Available online at arXiv:1412.0781 [cs.CV]}, 2014.

\end{thebibliography}

\end{document}